\documentclass[10pt,twocolumn,letterpaper]{article}

\usepackage{iccv}
\usepackage{times}
\usepackage{epsfig}
\usepackage{graphicx}
\usepackage{amsmath}
\usepackage{amssymb}
\usepackage{caption}
\usepackage{subcaption}
\usepackage{epstopdf}
\usepackage{multirow}
\usepackage{epigraph}
\usepackage{balance}
\usepackage{url}
\usepackage{colortbl}
\usepackage{array}
\newcolumntype{C}[1]{>{\centering\let\newline\\\arraybackslash\hspace{0pt}}m{#1}}

\definecolor{orange}{rgb}{1,0.9,0.65}
\definecolor{gr}{rgb}{0.9,1,0.6}
\definecolor{bl}{rgb}{0.9,0.8,1}
\definecolor{bg}{rgb}{0.8,0.9,1}
\definecolor{orr}{rgb}{1,0.85,0.4}
\definecolor{grr}{rgb}{0.8,1,0.5}
\definecolor{blr}{rgb}{0.85,0.45,1}
\definecolor{bgr}{rgb}{0.5,0.9,1}
\definecolor{gry}{rgb}{0.92,0.92,0.92}

\usepackage[pagebackref=true,breaklinks=true,letterpaper=true,colorlinks,bookmarks=false]{hyperref}

\iccvfinalcopy 

\def\iccvPaperID{1754} 
\def\httilde{\mbox{\tt\raisebox{-.5ex}{\symbol{126}}}}

\ificcvfinal\fi
\begin{document}

\title{Movie to Book Alignment via Video-to-Text and Text-to-Text Neural Embeddings}
\title{\vspace{-8mm}Aligning Books and Movies: Towards Story-like Visual Explanations by Watching Movies and Reading Books\\[-0.5mm]}

\author{Yukun Zhu~\thanks{Denotes equal contribution}$\,\,^{,1}$  \quad
Ryan Kiros\textsuperscript{*,1} \quad
Richard Zemel$^1$ \quad
Ruslan Salakhutdinov$^1$ \quad \\
Raquel Urtasun$^1$ \quad
Antonio Torralba$^2$ \quad
Sanja Fidler$^1$\\[0.5mm]
$^1$University of Toronto \quad
$^2$Massachusetts Institute of Technology\\
{\tt\small \{yukun,rkiros,zemel,rsalakhu,urtasun,fidler\}@cs.toronto.edu, torralba@csail.mit.edu}
}

\maketitle

\begin{abstract}
Books are a rich source of both fine-grained information, how a character, an object or a scene looks like, as well as high-level semantics, what someone is thinking, feeling and how these states evolve through a story. This paper aims to align books to their movie releases in order to provide rich descriptive explanations for visual content that go semantically far beyond the captions available in  current datasets. To align movies and books we exploit a neural sentence embedding that is trained in an unsupervised way from a large corpus of books, as well as a video-text neural embedding for computing similarities between movie clips and sentences in the book. We propose a context-aware CNN to combine information from multiple sources. We demonstrate good quantitative performance for movie/book alignment and show several qualitative examples that showcase the diversity of tasks our model can be used for.
\end{abstract}
\vspace{-4mm}

\section{Introduction}

A truly intelligent machine needs to not only parse the surrounding 3D environment, but also understand why people take certain actions, what they will do next, what they could possibly be thinking, and even try to  empathize with them. In this quest, language will play a crucial role in grounding visual information to high-level semantic concepts. Only a few words in a sentence may convey really rich semantic information. Language also represents a natural means of interaction between a naive user and our vision algorithms, which is particularly important for applications such as social robotics or assistive driving. 

Combining images or videos with language has gotten significant attention in the past year, partly due to the creation of  CoCo~\cite{lin2014microsoft}, Microsoft's large-scale captioned image dataset. The field has tackled a diverse set of tasks such as captioning~\cite{DBLP:journals/corr/KirosSZ14,Karpathy15,Vinyals14,Venugopalan:2014wc,Mao14}, alignment~\cite{Karpathy15,KongCVPR14,Tapaswi_J1_PlotRetrieval}, Q\&A~\cite{malinowski14nips,Lin15}, visual model learning from textual descriptions~\cite{Gupta08,ramanathan13}, and semantic visual search with natural multi-sentence queries~\cite{Lin:2014db}.

Books provide us with very rich, descriptive text that conveys both fine-grained visual details (how people or scenes look like) as well as high-level semantics (what people think and feel, and how their states evolve through a story). This source of knowledge, however, does not come with associated visual information that would enable us to ground it with descriptions. Grounding descriptions in books to vision would allow us to get textual \emph{explanations} or stories behind visual information rather than simplistic \emph{captions} available in current datasets. It can also provide us with extremely large amount of data (with tens of thousands books available online). 

In this paper, we exploit the fact that many books have been turned into movies. Books and their movie releases have a lot of common knowledge as well as they are complementary in many ways. For instance, books provide detailed descriptions about the intentions and mental states of the characters, while movies are better at capturing visual aspects of the settings. 

\begin{figure}[t!]
\vspace{-3mm}
\includegraphics[width=1\linewidth,trim = 0mm 26mm 8mm 0mm, clip=true]{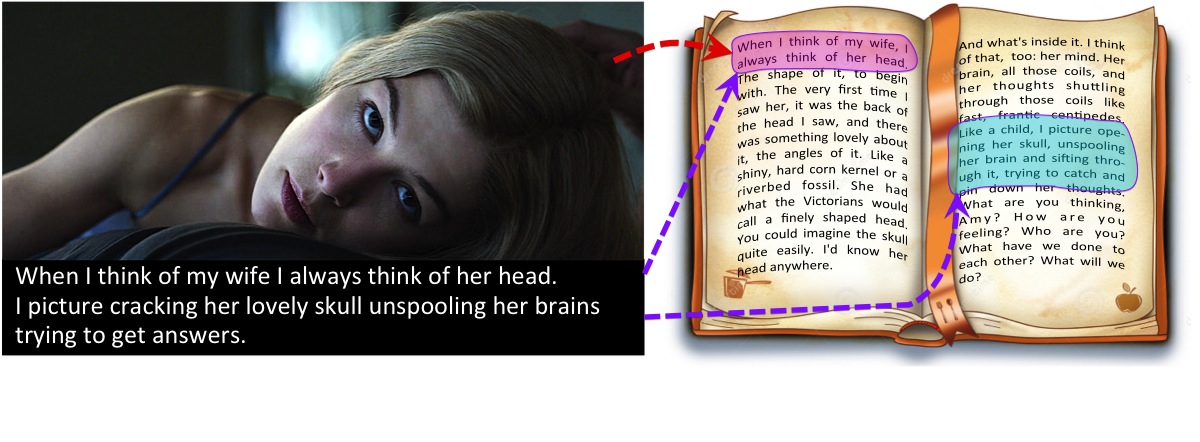}
\vspace{-7mm}
\caption{Shot from the movie \emph{Gone Girl}, along with the subtitle, aligned with the book. We reason about the visual and dialog (text) alignment between the movie and a book.}
\label{fig:intro}
\vspace{-2mm}
\end{figure}

The first challenge we need to address, and the focus of this paper, is to align books with their movie releases in order to obtain rich descriptions for the visual content. We aim to align the two sources with two types of information: \emph{visual}, where the goal is to link a movie shot to a book paragraph, and \emph{dialog}, where we want to find correspondences between sentences in the movie's subtitle and sentences in the book. We formulate the problem of movie/book alignment as finding correspondences between shots in the movie as well as dialog sentences in the subtitles and sentences in the book (Fig.~\ref{fig:intro}). We introduce a novel sentence similarity measure based on a neural sentence embedding trained on millions of sentences from a large corpus of books. On the visual side, we extend the neural image-sentence embeddings to the video domain and train the model on DVS descriptions of movie clips. Our approach combines different similarity measures and takes into account contextual information contained in the nearby shots and book sentences. Our final alignment model is formulated as an energy minimization problem that encourages the alignment to follow a similar timeline. To evaluate the book-movie alignment model we collected a dataset with 11 movie/book pairs annotated with 2,070 shot-to-sentence correspondences. We demonstrate good quantitative performance and show several qualitative examples that showcase the diversity of tasks our model can be used for. 

The alignment model can have multiple applications. Imagine an app which allows the user to browse the book as the scenes unroll in the movie: perhaps its ending or acting are ambiguous, and one would like to query the book for  answers.  Vice-versa, while reading the book one might want to switch from text to video, particularly for the juicy scenes. We also show other applications of learning from movies and books such as book retrieval (finding the book that goes with a movie and finding other similar books), and captioning CoCo images with story-like descriptions.

\begin{table*}[t]
\vspace{-3mm}
\centering
\scriptsize
\begin{tabular}{|l|C{0.7cm}|C{0.8cm}|C{0.85cm}|C{1.27cm}|C{1.27cm}|C{0.88cm}|C{0.69cm}|C{0.92cm}|C{0.8cm}|C{0.78cm}|}
\hline
 & \multicolumn{6}{>{\columncolor{bgr}}c|}{{\sc book}} & \multicolumn{2}{>{\columncolor{orr}}c|}{{\sc movie}} & \multicolumn{2}{>{\columncolor{gr}}c|}{{\sc annotation}}\\ 
\hline
Title & \# sent. & \# words & \# unique words & avg. \# words per sent. & max \# words per sent. & \# paragraphs & \# shots & \# sent. in subtitles & \# dialog align. & \# visual align.\\ 
\hline
Gone Girl & 12,603 & 148,340 & 3,849 & 15 & 153 & 3,927 & 2,604 & 2,555 & 76 & 106 \\ 
\hline
Fight Club & 4,229 & 48,946 & 1,833 & 14 & 90 & 2,082 & 2,365 & 1,864 & 104 & 42 \\ 
\hline
No Country for Old Men & 8,050 & 69,824 & 1,704 & 10 & 68 & 3,189 & 1,348 & 889 & 223 & 47 \\ 
\hline
Harry Potter and the Sorcerers Stone & 6,458 & 78,596 & 2,363 & 15 & 227 & 2,925 & 2,647 & 1,227 & 164 & 73 \\ 
\hline
Shawshank Redemption & 2,562 & 40,140 & 1,360 & 18 & 115 & 637 & 1,252 & 1,879 & 44 & 12 \\ 
\hline
The Green Mile & 9,467 & 133,241 & 3,043 & 17 & 119 & 2,760 & 2,350 & 1,846 & 208 & 102 \\ 
\hline
American Psycho & 11,992 & 143,631 & 4,632 & 16 & 422 & 3,945 & 1,012 & 1,311 & 278 & 85 \\ 
\hline
One Flew Over the Cuckoo Nest & 7,103 & 112,978 & 2,949 & 19 & 192 & 2,236 & 1,671 & 1,553 & 64 & 25 \\ 
\hline
The Firm & 15,498 & 135,529 & 3,685 & 11 & 85 & 5,223 & 2,423 & 1,775 & 82 & 60 \\ 
\hline
Brokeback Mountain & 638 & 10,640 & 470 & 20 & 173 & 167 & 1,205 & 1,228 & 80 & 20 \\ 
\hline
The Road & 6,638 & 58,793 & 1,580 & 10 & 74 & 2,345 & 1,108 & 782 & 126 & 49 \\ 
\hline
\hline
{\bf All} & 85,238 & 980,658 & 9,032 & 15 & 156 & 29,436 & 19,985 & 16,909 & 1,449 & 621 \\
\hline
\end{tabular}
\vspace{-2.5mm}
\caption{Statistics for our {\bf MovieBook Dataset} with ground-truth for alignment between books and their movie releases. }
\label{tab:datasetstats}
\vspace{-2mm}
\end{table*}

\begin{table*}
\small
\begin{center}
\mbox{
\begin{tabular}{| c | c | c | c | c | c |}
\hline
\# of books & \# of sentences & \# of words & \# of unique words & mean \# of words per sentence & median \# of words per sentence \\
\hline
11,038 & 74,004,228 & 984,846,357 & 1,316,420 & 13 & 11 \\
\hline
\end{tabular}}
\end{center}
\vspace{-6mm}
\caption{Summary statistics of our {\bf BookCorpus} dataset. We use this corpus to train the sentence embedding model.}
\label{tab:bookdata}
\vspace{-3mm}
\end{table*}

\vspace{-3mm}
\section{Related Work} \label{sec:rw}

Most effort in the domain of vision and language has been devoted to the problem of image captioning. Older work made use of fixed visual representations 
and translated them into textual descriptions~\cite{Farhadi10,BabyTalk}. Recently, several approaches based on RNNs emerged, generating captions via a learned joint image-text embedding~\cite{DBLP:journals/corr/KirosSZ14,Karpathy15,Vinyals14,Mao14}. These approaches have also been extended to generate descriptions of short video clips~\cite{Venugopalan:2014wc}. In~\cite{thewhy}, the authors  go beyond describing \emph{what} is happening in an image and provide explanations about \emph{why} something is happening. 

For text-to-image alignment,~\cite{KongCVPR14,FidlerCVPR13} find correspondences between nouns and pronouns in a caption and visual objects using several visual and textual potentials. Lin~\etal~\cite{Lin:2014db} does so for videos. In~\cite{Karpathy15}, the authors use RNN embeddings to find the correspondences. ~\cite{Xu15} combines neural embeddings with soft attention in order to align the words to image regions. 

Early work on movie-to-text alignment include dynamic time warping for aligning movies to scripts with the help of subtitles~\cite{Everingham:2006jr,Cour08}. Sankar~\etal~\cite{Sankar:bv} further developed a system which identified sets of visual  and audio features to align movies and scripts without making use of the subtitles. Such alignment has been exploited to provide weak labels for person naming tasks~\cite{Everingham:2006jr,Sivic:2009kt,Ramanathan:2014fj}. 

Closest to our work is~\cite{Tapaswi_J1_PlotRetrieval}, which aligns plot synopses to shots in the TV series for story-based content retrieval. This work adopts a similarity function between sentences in plot synopses and shots based on person identities and keywords in subtitles. Our work differs with theirs in several important aspects. First, we tackle a more challenging problem of movie/book alignment. Unlike plot synopsis, which closely follow the storyline of movies, books are more verbose and might vary in the storyline from their movie release. Furthermore, we use learned neural embeddings to compute the similarities rather than hand-designed similarity functions.

Parallel to our work,~\cite{Tapaswi2015_Book2Movie} aims to align scenes in movies to chapters in the book. However, their approach operates on a very coarse level (chapters), while ours does so on the sentence/paragraph level. Their dataset thus evaluates on 90 scene-chapter correspondences, while our dataset draws 2,070 shot-to-sentences alignments. Furthermore, the approaches are inherently different. ~\cite{Tapaswi2015_Book2Movie} matches the presence of characters in a scene to those in a chapter, as well as uses hand-crafted similarity measures between sentences in the subtitles and dialogs in the books, similarly to~\cite{Tapaswi_J1_PlotRetrieval}. 

Rohrbach~\etal~\cite{Rohrbach15} recently released the Movie Description dataset which contains clips from movies, each time-stamped with a sentence from DVS (Descriptive Video Service). The dataset contains clips from over a 100 movies, and provides a great resource for the captioning techniques. Our effort here is to align movies with books in order to obtain longer, richer and more high-level video descriptions.

We start by describing our new dataset, and then explain our proposed approach. 

\section{The MovieBook and BookCorpus Datasets}

We collected two large datasets, one for movie/book alignment and one with a large number of books. 

\vspace{-4mm}
\paragraph{The MovieBook Dataset.} Since no prior work or data exist on the problem of movie/book alignment, we collected a new dataset with $11$ movies along with the books on which they were based on. For each movie we also have a subtitle file, which we parse into a set of time-stamped sentences. Note that no speaker information is provided in the subtitles. We automatically parse each book into sentences, paragraphs (based on indentation in the book),  and chapters (we assume a chapter title has indentation, starts on a new page, and does not end with an end symbol). 

Our annotators had the movie and a book opened side by side. They were asked to iterate between browsing the book and watching a few shots/scenes of the movie, and trying to find correspondences between them. In particular, they marked the exact time (in seconds) of correspondence in the movie and the matching  line number in the book file, indicating the beginning of the matched sentence. On the video side, we assume that the match spans across a \emph{shot} (a video unit with smooth camera motion). If the match was longer in duration, the annotator also indicated the ending time of the match. Similarly for the book, if more sentences matched, the annotator indicated from which to which line a match occurred. Each alignment was also tagged, indicating whether it was a \textit{visual}, \textit{dialogue}, or an \textit{audio match}. Note that even for dialogs, the movie and book versions are semantically similar but not exactly the same. Thus deciding on what defines a match or not is also somewhat subjective and may slightly vary across our annotators. 
Altogether, the annotators spent $90$ hours labeling $11$ movie/book pairs, locating  2,070 correspondences. 

Table~\ref{tab:datasetstats} presents our dataset, while Fig.~\ref{fig:align} shows a few ground-truth alignments. One can see the complexity and diversity of the data: the number of sentences per book vary from 638 to 15,498, even though the movies are similar in duration. This indicates a huge diversity in descriptiveness across literature, and presents a challenge for matching. The sentences also vary in length, with the sentences in Brokeback Mountain being twice as long as those in The Road. The longest sentence in American Psycho has 422 words and spans over a page in the book.

Aligning movies with books is challenging even for humans, mostly due to the scale of the data. Each movie is on average  2h long and has 1,800 shots, while a book has on average 7,750 sentences. Books also have different styles of writing, formatting, different and challenging language, slang (\emph{going} vs \emph{goin'}, or even \emph{was} vs \emph{'us}), etc. 
As one can see from Table~\ref{tab:datasetstats}, finding visual matches turned out to be particularly challenging. This is because the visual descriptions in books can be either very short and hidden within longer paragraphs or even within a longer sentence, or very verbose -- in which case they get obscured with the surrounding text -- and are hard to spot. Of course, how close the movie follows the book is also up to the director, which can be seen through the number of alignments that our annotators found across different movie/books. 

\vspace{-4mm}
\paragraph{The BookCorpus Dataset.} In order to train our sentence similarity model we collected a corpus of 11,038 books from the web. These are free books written by yet unpublished authors. We only included books that had more than 20K words in order to filter out perhaps noisier shorter stories. The dataset has books in 16 different genres, e.g., \emph{Romance} (2,865 books), \emph{Fantasy} (1,479), \emph{Science fiction} (786), \emph{Teen} (430), etc.  Table~\ref{tab:bookdata} highlights the summary statistics of our book corpus.
\section{Aligning Books and Movies}

Our approach aims to align a movie with a book by exploiting visual information as well as dialogs. We take shots as video units  and sentences from subtitles to represent dialogs. Our goal is to match these to the sentences in the book. We propose several measures to compute similarities between pairs of sentences as well as shots and sentences. We use our novel  deep neural embedding trained on our large corpus of books to predict similarities between sentences. Note that an extended version of the sentence embedding is  described in detail in~\cite{skipthoughts} showing how to deal with million-word vocabularies, and demonstrating its performance on a large variety of NLP benchmarks. 
For comparing shots with sentences we extend the neural embedding of images and text~\cite{DBLP:journals/corr/KirosSZ14} to operate in the video domain. We next develop a novel contextual alignment model that combines information from various similarity measures and a larger time-scale in order to make better local alignment predictions. Finally, we propose a simple pairwise Conditional Random Field (CRF) that smooths the alignments by encouraging them to follow a linear timeline, both in the video and book domain.  

We first explain our sentence, followed by our joint video to text embedding. We next propose our contextual model that combines  similarities and discuss CRF in more detail. 

\subsection{Skip-Thought Vectors}

\begin{figure*}[tp]
\vspace{-6mm}
    \label{figure:bookmodel}
    \centering
        \vspace{2.5mm}
        \includegraphics[width=0.86\textwidth]{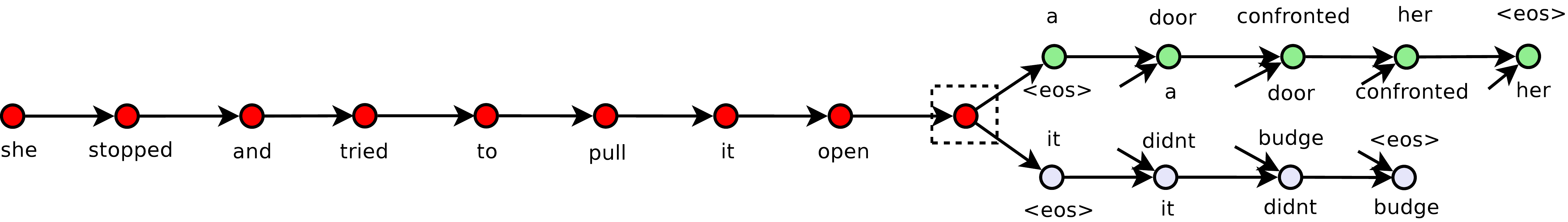}
        \vspace{-1mm}
        \caption{\label{fig:skip-gram} Sentence neural embedding~\cite{skipthoughts}. Given a tuple ($s_{i-1}, s_i, s_{i+1}$) of consecutive sentences in text, where $s_i$ is the $i$-th sentence, we encode $s_i$ and aim to reconstruct the previous $s_{i-1}$ and the following sentence $s_{i+1}$. Unattached arrows are connected to the encoder output. Colors depict which components share parameters. $\langle$eos$\rangle$ is the end of sentence token. }
        \vspace{-1.5mm}
\end{figure*}

\begin{table*}
\footnotesize
\begin{center}
\mbox{
\addtolength{\tabcolsep}{3pt}
\begin{tabular}{| c | c |}
\hline

& he started the car , left the parking lot and merged onto the highway a few miles down the road .  \\
 & he shut the door and watched the taxi drive off . \\
he drove down the street off  into the distance . & she watched the lights flicker through the trees as the men drove toward the road .  \\
& he jogged down the stairs , through the small lobby , through the door and into the street . \\
\hline
& a messy business to be sure , but necessary to achieve a fine and noble end .  \\
 & they saw their only goal as survival and logically planned a strategy to achieve it . \\
the most effective way to end the battle . & there would be far fewer casualties and far less destruction .  \\
 & the outcome was the lisbon treaty . \\
\hline
\end{tabular}}
\end{center}
\vspace{-5mm}
\caption{Qualitative results from the sentence embedding model. For each query sentence on the left, we retrieve the 4 nearest neighbor sentences (by inner product) chosen from books the model has not seen before.}
\label{tab:knnsents}
\vspace{-3mm}
\end{table*}

In order to score the similarity between two sentences, we exploit our architecture for learning unsupervised representations of text~\cite{skipthoughts}. The model is loosely inspired by the skip-gram~\cite{mikolov2013efficient} architecture for learning representations of words. In the word skip-gram model, a word $w_i$ is chosen and must predict its surrounding context (e.g. $w_{i+1}$ and $w_{i-1}$ for a context window of size 1). Our model works in a similar way but at the sentence level. That is, given a sentence tuple ($s_{i-1}, s_i, s_{i+1}$) our model first encodes the sentence $s_i$ into a fixed vector, then conditioned on this vector tries to reconstruct the sentences $s_{i-1}$ and $s_{i+1}$, as shown in Fig.~\ref{fig:skip-gram}. The motivation for this architecture is inspired by the distributional hypothesis: sentences that have similar surrounding context are likely to be both semantically and syntactically similar. Thus, two sentences that have similar syntax and semantics are likely to be encoded to a similar vector. Once the model is trained, we can map any sentence through the encoder to obtain vector representations, then score their similarity through an inner product.

The learning signal of the model depends on having contiguous text, where sentences follow one another in sequence. A natural corpus for training our model is thus a large collection of books. 
Given the size and diversity of genres, our BookCorpus allows us to learn very general representations of text. For instance, Table~\ref{tab:knnsents} illustrates the nearest neighbours of query sentences, taken from held out books that the model was not trained on. These qualitative results demonstrate that our intuition is correct, with resulting nearest neighbors corresponds largely to syntactically and semantically similar sentences. Note that the sentence embedding  is general and can be applied to  other domains not considered in this paper, which is explored in~\cite{skipthoughts}. 

To construct an encoder, we use a recurrent neural network, inspired by the success of encoder-decoder models for neural machine translation \cite{kalchbrenner2013recurrent, cho2014learning, bahdanau2014neural, sutskever2014sequence}. Two kinds of activation functions have recently gained traction: long short-term memory (LSTM) \cite{hochreiter1997long} and the gated recurrent unit (GRU) \cite{chung2014empirical}. Both types of activation successfully solve the vanishing gradient problem, through the use of gates to control the flow of information. The LSTM unit explicity employs a cell that acts as a carousel with an identity weight. The flow of information through a cell is controlled by input, output and forget gates which control what goes into a cell, what leaves a cell and whether to reset the contents of the cell. The GRU does not use a cell but employs two gates: an update and a reset gate. In a GRU, the hidden state is a linear combination of the previous hidden state and the proposed hidden state, where the combination weights are controlled by the update gate. GRUs have been shown to perform just as well as LSTM on several sequence prediction tasks~\cite{chung2014empirical} while being simpler. Thus, we use GRU as the activation function for our encoder and decoder RNNs.

Suppose that we are given a sentence tuple ($s_{i-1}, s_i, s_{i+1}$), and let $w^t_i$ denote the $t$-th word for $s_i$ and let ${\bf x}^t_i$  be its word embedding. We break the model description into three parts: the encoder, decoder and objective function.

\noindent {\bf Encoder.} Let $w^1_i, \ldots, w^N_i$ denote words in sentence $s_i$ with $N$ the number of words in the sentence. The encoder produces a hidden state ${\bf h}^t_i$ at each time step which forms the representation of the sequence $w^1_i, \ldots, w^t_i$. Thus, the hidden state ${\bf h}^N_i$ is the representation of the whole sentence. The GRU produces the next hidden state as a linear combination of the previous hidden state and the proposed state update (we drop subscript $i$):
\begin{equation}
{\bf h}^t = (1 - {\bf z}^t) \odot {\bf h}^{t-1} + {\bf z}^t \odot \bar{{\bf h}}^t
\end{equation}
where $\bar{{\bf h}}^t$ is the proposed state update at time $t$, ${\bf z}^t$ is the update gate and ($\odot$) denotes a component-wise product. The update gate takes values between zero and one. In the extreme cases, if the update gate is the vector of ones, the previous hidden state is completely forgotten and ${\bf h}^t = \bar{{\bf h}}^t$. Alternatively, if the update gate is the zero vector, than the hidden state from the previous time step is simply copied over, that is ${\bf h}^t = {\bf h}^{t-1}$. The update gate is computed as
\begin{equation}
{\bf z}^t = \sigma ( {\bf W}_z {\bf x}^t + {\bf U}_z {\bf h}^{t-1} )
\end{equation}
where ${\bf W}_z$ and ${\bf U}_z$ are the update gate parameters. The proposed state update is given by
\begin{equation}
\bar{{\bf h}}^t = \text{tanh} ( {\bf W} {\bf x}^t + {\bf U} ( {\bf r}^t \odot {\bf h}^{t-1} ))
\end{equation}
where ${\bf r}_t$ is the reset gate, which is computed as
\begin{equation}
{\bf r}^t = \sigma ( {\bf W}_r {\bf x}^t + {\bf U}_r {\bf h}^{t-1} )
\end{equation}
If the reset gate is the zero vector, than the proposed state update is computed only as a function of the current word. Thus after iterating this equation sequence for each word, we obtain a sentence vector ${\bf h}^N_i = {\bf h}_i$ for sentence $s_i$.   

\noindent {\bf Decoder.} The decoder computation is analogous to the encoder, except that the computation is conditioned on the sentence vector ${\bf h}_i$. Two separate decoders are used, one for the previous sentence $s_{i-1}$ and one for the next sentence $s_{i+1}$. These decoders use different parameters to compute their hidden states but both share the same vocabulary matrix ${\bf V}$ that takes a hidden state and computes a distribution over words. Thus, the decoders are analogous to an RNN language model but conditioned on the encoder sequence. Alternatively, in the context of image caption generation, the encoded sentence ${\bf h}_i$ plays a similar role as the image.

We describe the decoder for the next sentence $s_{i+1}$ (computation for $s_{i-1}$ is identical). Let ${\bf h}^t_{i+1}$ denote the hidden state of the decoder at time $t$. The update and reset gates for the decoder are given as follows (we drop $i+1$):
\begin{eqnarray}
{\bf z}^t &=& \sigma ( {\bf W}_z^d {\bf x}^{t-1} + {\bf U}_z^d {\bf h}^{t-1} + {\bf C}_z {\bf h}_i ) \\
{\bf r}^t &=& \sigma ( {\bf W}_r^d {\bf x}^{t-1} + {\bf U}_r^d {\bf h}^{t-1} + {\bf C}_r {\bf h}_i )
\end{eqnarray}
the hidden state ${\bf h}^t_{i+1}$ is then computed as:
\begin{eqnarray}
& & \bar{{\bf h}}^t = \text{tanh} ( {\bf W}^d {\bf x}^{t-1} + {\bf U}^d ( {\bf r}^t \odot {\bf h}^{t-1} ) + {\bf C} {\bf h}_i) \\
& & {\bf h}^t_{i+1} = (1 - {\bf z}^t) \odot {\bf h}^{t-1} + {\bf z}^t \odot \bar{{\bf h}}^t
\end{eqnarray}
Given ${\bf h}^t_{i+1}$, the probability of word $w_{i+1}^t$ given the previous $t-1$ words and the encoder vector is
\begin{equation}
P(w_{i+1}^t | w_{i+1}^{<t}, {\bf h}_i) \propto \text{exp} ( {\bf v}_{w_{i+1}^t} {\bf h}^t_{i+1} )
\end{equation}
where ${\bf v}_{w_{i+1}^t}$ denotes the row of ${\bf V}$ corresponding to the word of $w_{i+1}^t$. An analogous computation is performed for the previous sentence $s_{i-1}$.

\noindent {\bf Objective.} Given ($s_{i-1}, s_i, s_{i+1}$), the objective optimized is the sum of log-probabilities for the next and previous sentences conditioned on the representation of the encoder:
\begin{equation}
\sum_t \text{log} P(w_{i+1}^t | w_{i+1}^{<t}, {\bf h}_i) + \sum_t \text{log} P(w_{i-1}^t | w_{i-1}^{<t}, {\bf h}_i)
\end{equation}
The total objective is the above summed over all such training tuples. Adam algorithm~\cite{kingma2014adam} is used for optimization.

\subsection{Visual-semantic embeddings of clips and DVS}

The model above describes how to obtain a similarity score between two sentences, whose representations are learned from millions of sentences in books. We now discuss how to obtain similarities between shots and sentences. 

Our approach closely follows the image-sentence ranking model proposed by \cite{DBLP:journals/corr/KirosSZ14}. In their model, an LSTM is used for encoding a sentence into a fixed vector. A linear mapping is applied to image features from a convolutional network. 
A score is computed based on the inner product between the normalized sentence and image vectors. Correct image-sentence pairs are trained to have high score, while incorrect pairs are assigned low scores.

In our case, we learn a visual-semantic embedding between movie clips and their DVS description. DVS (``Descriptive Video Service'') is a service that inserts audio descriptions of the movie between the dialogs in order to enable the visually impaired to follow the movie like anyone else. We used the movie description dataset of~\cite{Rohrbach15} for learning our embedding. This dataset has $94$ movies, and 54,000 described clips. We represent each movie clip as a vector corresponding to mean-pooled features across each frame in the clip. We used the GoogLeNet architecture~\cite{szegedy2014going} as well as hybrid-CNN~\cite{ZhouNIPS2014} for extracting frame features. For DVS, we pre-processed the descriptions by removing names and replacing these with a {\it someone} token. 

The LSTM architecture in this work is implemented using the following equations. As before, we represent a word embedding at time $t$ of a sentence as ${\bf x}^t$:
\vspace{-2mm}
\begin{eqnarray}
{\bf i}^t &=& \sigma( {\bf W}_{xi} {\bf x}^t + {\bf W}_{hi} {\bf m}^{t-1} + {\bf W}_{ci} {\bf c}^{t-1} ) \\
{\bf f}^t &=& \sigma( {\bf W}_{xf} {\bf x}^t + {\bf W}_{hf} {\bf m}^{t-1} + {\bf W}_{cf} {\bf c}^{t-1} )\\
{\bf a}^t &=& \text{tanh} ( {\bf W}_{xc} {\bf x}^t + {\bf W}_{hc} {\bf m}^{t-1} ) \\
{\bf c}^t &=& {\bf f}^t \odot {\bf c}^{t-1} + {\bf i}^t \odot {\bf a}^t \\
{\bf o}^t &=& \sigma( {\bf W}_{xo} {\bf x}^t + {\bf W}_{ho} {\bf m}^{t-1} + {\bf W}_{co} {\bf c}^t ) \\
{\bf m}^t &=& {\bf o}^t \odot \text{tanh} ({\bf c}^t)
\end{eqnarray}
where ($\sigma$) denotes the sigmoid activation function and ($\odot$) indicates component-wise multiplication. The states (${\bf i}^t, {\bf f}^t, {\bf c}^t, {\bf o}^t, {\bf m}^t$) correspond to the input, forget, cell, output and memory vectors, respectively. If the sentence is of length $N$, then the vector ${\bf m}^N = {\bf m}$ is the vector representation of the sentence.

Let ${\bf q}$ denote a movie clip vector, and let ${\bf v} = {\bf W}_I {\bf q}$ be the embedding of the movie clip. We define a scoring function $s({\bf m}, {\bf v}) = {\bf m} \cdot {\bf v}$, where ${\bf m}$ and ${\bf v}$ are first scaled to have unit norm (making $s$ equivalent to cosine similarity). We then optimize the following pairwise ranking loss:
\vspace{-2mm}
\begin{eqnarray}
\underset{\boldsymbol\theta}{\operatorname{min}} \hspace{1mm}
\label{eq:rank}
\sum_{\bf m} \sum_k \text{max}\{0, \alpha - s({\bf m}, {\bf v}) + s({\bf m}, {\bf v}_k)\} \\
+ \sum_{\bf v} \sum_k \text{max}\{0, \alpha - s({\bf v}, {\bf m}) + s({\bf v}, {\bf m}_k)\}, 
\nonumber 
\end{eqnarray}
with ${\bf m}_k$ a contrastive (non-descriptive) sentence vector for a clip embedding ${\bf v}$, and vice-versa with ${\bf v}_k$. We train our model with stochastic gradient descent without momentum.

\subsection{Context aware similarity}
\label{sec:context}

We employ the clip-sentence embedding to compute similarities between each shot in the movie and each sentence in the book. For dialogs, we use several  similarity measures each capturing a different level of semantic similarity. We compute BLEU~\cite{BLEU} between each subtitle and book sentence to identify nearly identical matches. Similarly to~\cite{Tapaswi_J1_PlotRetrieval}, we use a tf-idf measure to find near duplicates but weighing down the influence of the less frequent words. Finally, we use our sentence embedding learned from books to score pairs of sentences that are semantically similar but may have a very different wording (i.e., paraphrasing). 

These similarity measures indicate the alignment between the two modalities. However, at the local, sentence level, alignment  can be rather ambiguous. For example, despite being a rather dark book, \emph{Gone Girl} contains $15$ occurrences of the sentence ``I love you''. We exploit the fact that a match is not completely isolated but that the sentences (or shots) around it are also to some extent similar.

\begin{figure}[t!]
\vspace{-2mm}
\centering
\includegraphics[width=0.75\linewidth]{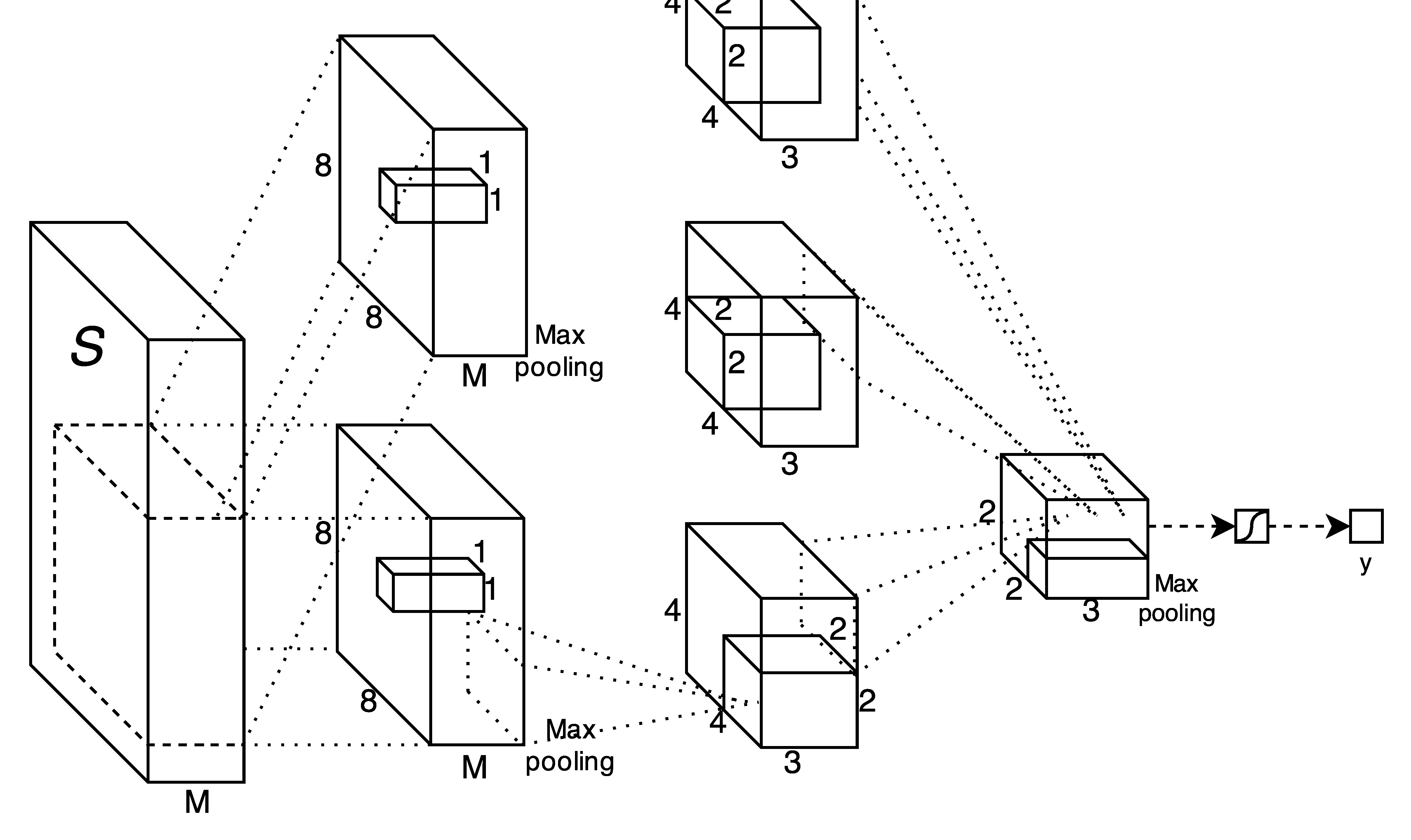}
\vspace{-4mm}
\caption{Our CNN for context-aware similarity computation. It has $3$ conv. layers and a sigmoid layer on top.}\label{fig:cnn}
\vspace{-2mm}
\end{figure}

We design a context aware similarity measure that takes into account all individual similarity measures as well as a fixed context window in both, the movie and book domain, and predicts a new similarity score. We stack a set of $M$ similarity measures into a tensor $S(i,j,m)$, where $i$, $j$, and $m$ are the indices of sentences in the subtitle, in the book, and individual similarity measures, respectively. In particular, we use $M=9$ similarities: visual and sentence embedding,  BLEU1-5, tf-idf, and a uniform prior. 
We want to predict a combined score $\mathrm{score}(i,j)=f(S(\mathbf{I,J,M}))$ at each location $(i,j)$ based on all measurements in a fixed volume defined by ${\bf I}$ around $i$, ${\bf J}$ around $j$, and $1,\dots,M$.  Evaluating the function $f(\cdot)$ at each location $(i,j)$ on a 3-D tensor $S$ is very similar to applying a convolution using a kernel of appropriate size. This motivates us to formulate the function $f(\cdot)$ as a deep convolutional neural network (CNN). In this paper, we adopt a 3-layer CNN as illustrated in Figure~\ref{fig:cnn}. We adopt the ReLU non-linearity with dropout to regularize our model. We optimize the cross-entropy loss over the training set using Adam algorithm. 


\subsection{Global Movie/Book Alignment}

So far, each shot/sentence was matched independently. However, most shots in movies and passages in the books follow a similar timeline. We would like to incorporate this prior into our alignment. In~\cite{Tapaswi_J1_PlotRetrieval}, the authors use dynamic time warping by enforcing that the shots in the movie can only match forward in time (to plot synopses in their case). 
However, the storyline of the movie and book can have crossings in time (Fig.~\ref{fig:align}), and the alignment might contain giant leaps forwards or backwards. Therefore, we formulate a movie/book alignment problem as inference in a Conditional Random Field that encourages nearby shots/dialog alignments to be consistent. Each node $y_i$ in our CRF represents an alignment of the shot in the movie with its corresponding subtitle sentence to a sentence in the book. Its state space is thus the set of all sentences in the book. 
The CRF energy of a configuration ${\bf y}$ is formulated as:
\vspace{-1mm}
\begin{align}\label{eqn:crf}
-\log p({\bf x},&{\bf y};\omega) = \sum_{i=1}^K \omega_u \phi_{u}(y_i) + \sum_{i=1}^{K}\sum_{j \in \mathcal{N}(i)} \omega_p\psi_{p}(y_i,y_j)\nonumber
\vspace{-2mm}
\end{align}
where $K$ is the number of nodes (shots), and $\mathcal{N}(i)$ the left and right neighbor of $y_i$.  Here, $\phi_u(\cdot)$ and $\psi_p(\cdot)$ are unary and pairwise potentials, respectively, and ${\bf \omega}=(\omega_u,\omega_p)$. We directly use the output of the CNN from~\ref{sec:context} as the unary potential $\phi_u(\cdot)$. For the pairwise potential, we measure the time span $d_s(y_i,y_j)$ between two neighbouring sentences in the subtitle and the distance $d_b(y_i,y_j)$ of their state space in the book. One pairwise potential is defined as: 
\begin{align}
\psi_{p}(y_i,y_j)=\frac{\left(d_s(y_i,y_j)-d_b(y_i,y_j)\right)^2}{\left(d_s(y_i,y_j)-d_b(y_i,y_j)\right)^2+\sigma^2}
\end{align}
Here $\sigma^2$ is a robustness parameter to avoid punishing giant leaps too harsh. Both $d_s$ and $d_b$ are normalized to $[0,\,1]$. In addition, we also employ another pairwise potential $\psi_{q}(y_i,y_j) = \frac{\left(d_b(y_i,y_j)\right)^2}{\left(d_b(y_i,y_j)\right)^2+\sigma^2}$ to encourage state consistency between nearby nodes. This potential is helpful when there is a long silence (no dialog) in the movie.    

\vspace{-3mm}
\paragraph{Inference.} Our CRF is a chain, thus  exact inference is possible using dynamic programming. We also prune some states that are very far from the uniform alignment (over $1/3$ length of the book) to further speed up computation.

\vspace{-3mm}
\paragraph{Learning.} Since ground-truth is only available for a sparse set of shots, we regard the states of unobserved nodes as hidden variables and learn the CRF weights with~\cite{Schwing12}. 

\vspace{-1mm}
\section{Experimental Evaluation}

We evaluate our model on our dataset of $11$ movie/book pairs. We train the parameters in our model (CNN and CRF) on \emph{Gone Girl}, and test our performance on the remaining $10$ movies. In terms of training speed, our video-text model ``watches'' 1,440 movies per day and our sentence model reads 870 books per day. We also show various qualitative results demonstrating the power of our approach. We provide more results in the Appendix of the paper.

\begin{figure*}[t!]
\vspace{-3mm}
\includegraphics[width=0.496\linewidth,trim=35 30 22 75,clip]{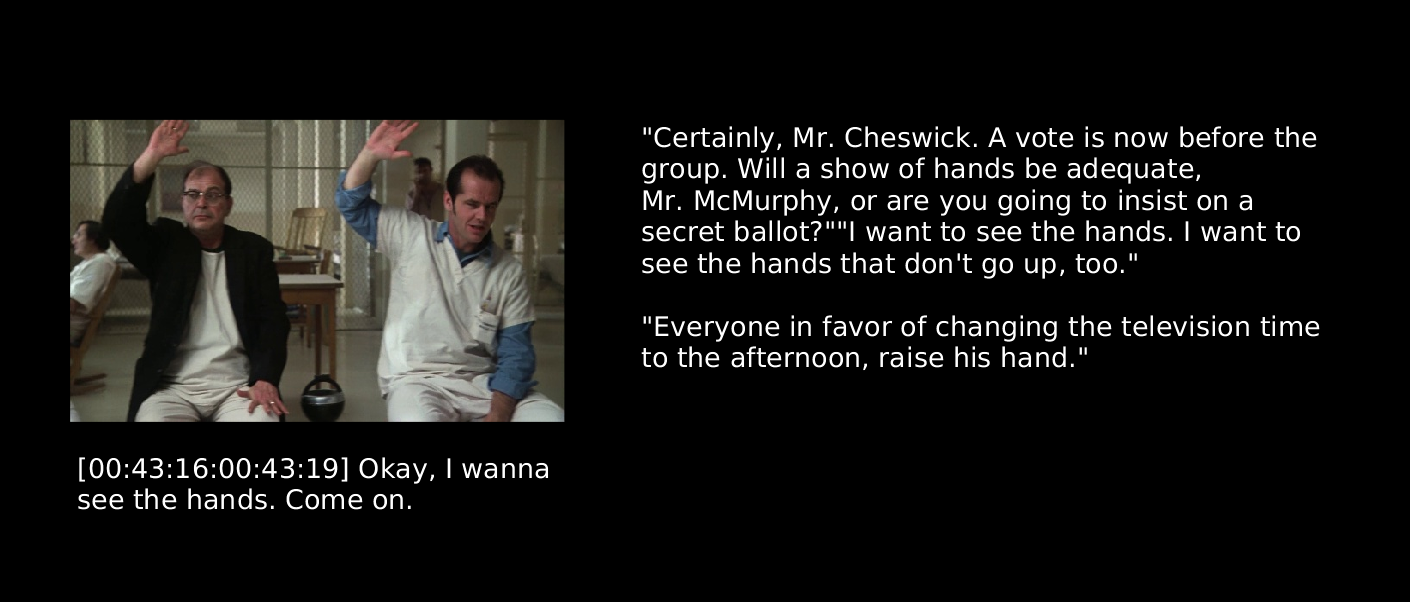}
\includegraphics[width=0.496\linewidth,trim=5 30 5 80,clip]{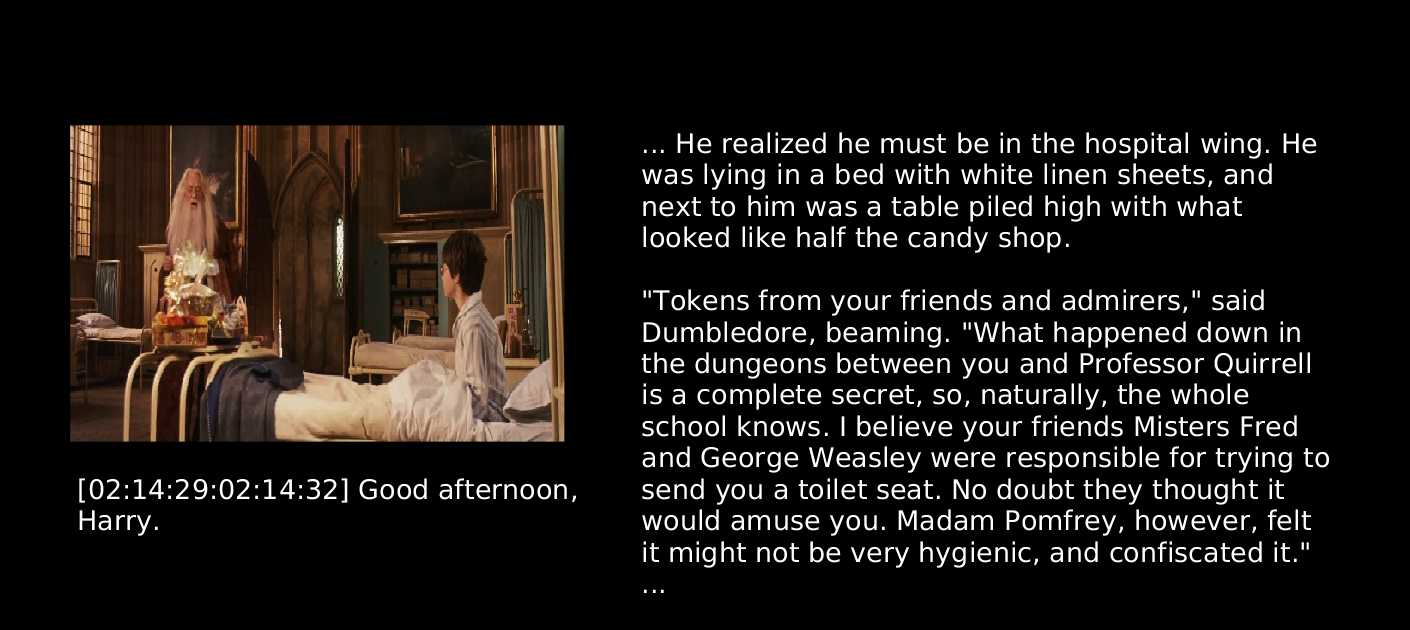}\\
\includegraphics[width=0.496\linewidth,trim=35 47 22 75,clip]{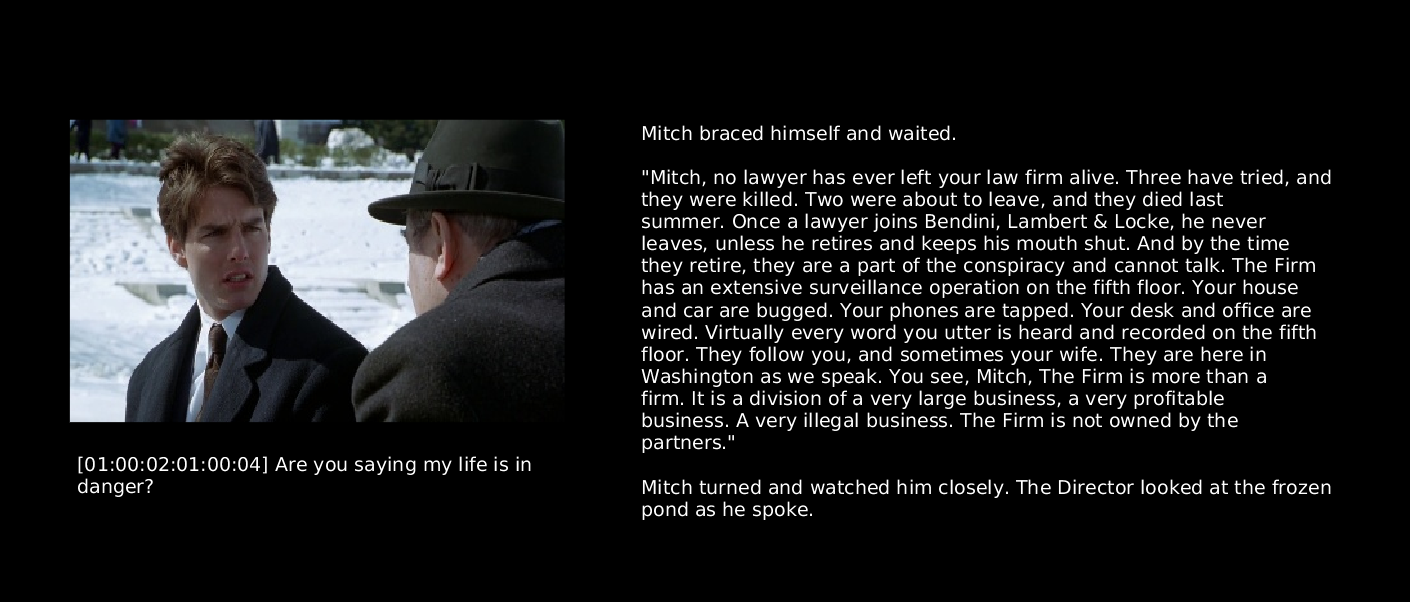}
\includegraphics[width=0.496\linewidth,trim=35 47 22 75,clip]{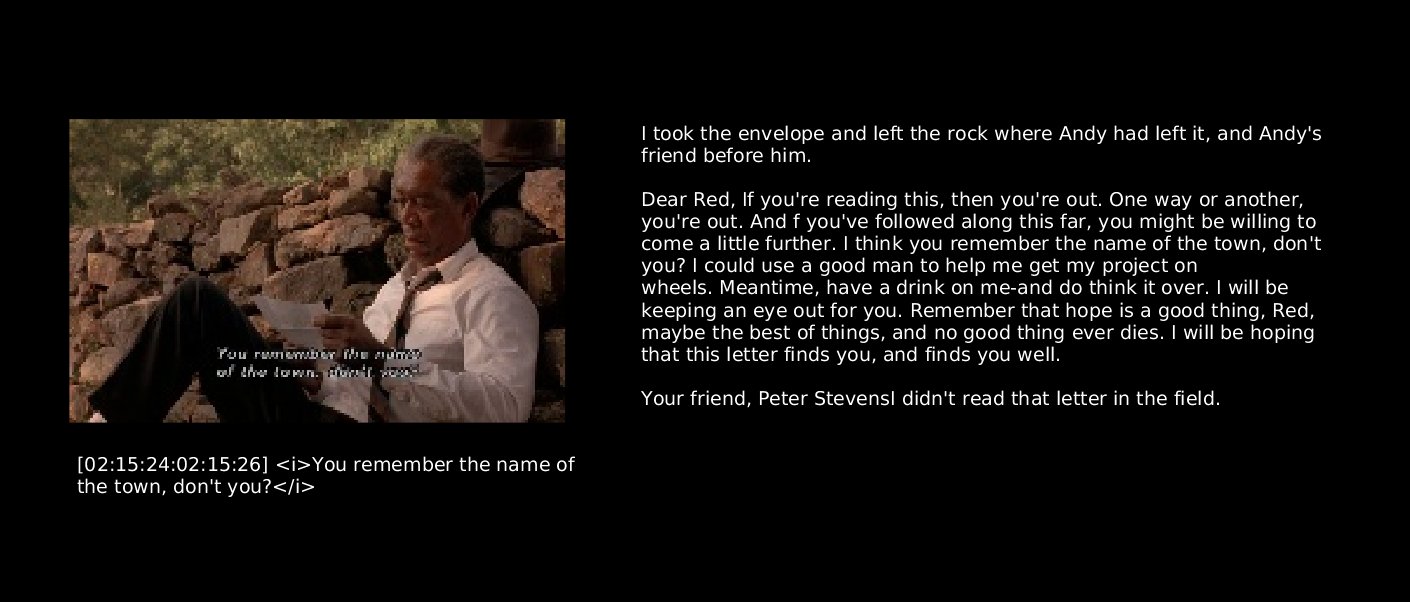}\\
\includegraphics[width=0.496\linewidth,trim=35 14 22 75,clip]{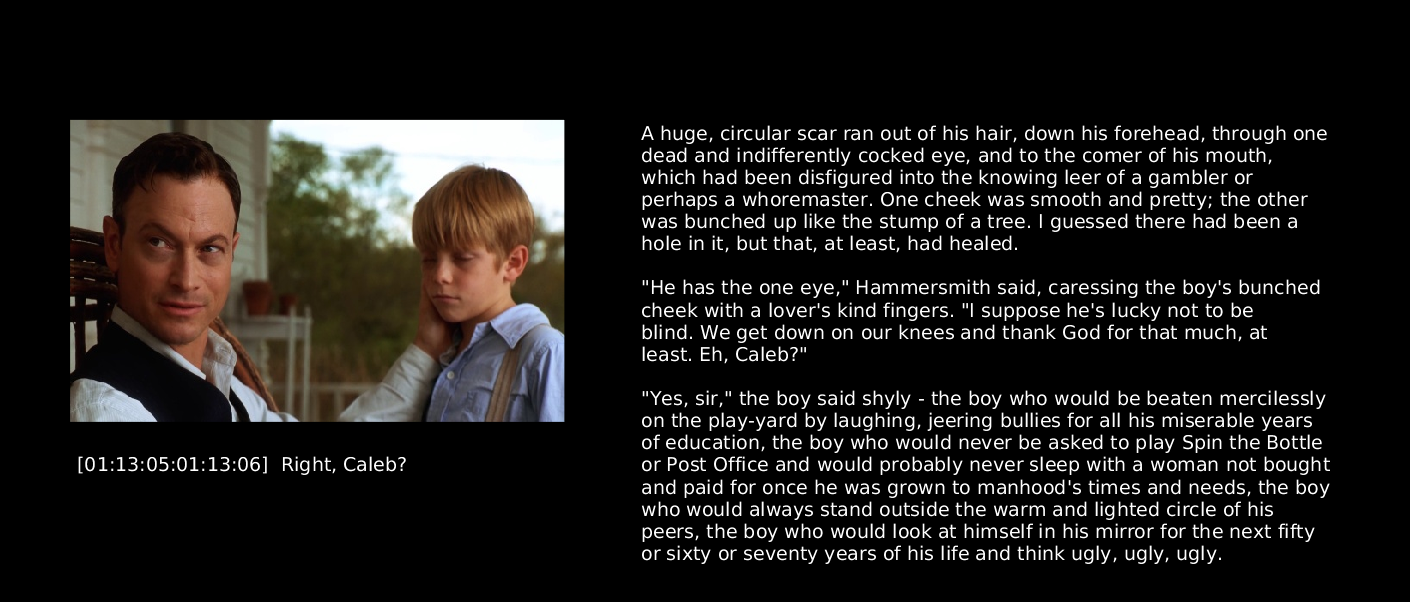}
\includegraphics[width=0.496\linewidth,trim=35 14 22 75,clip]{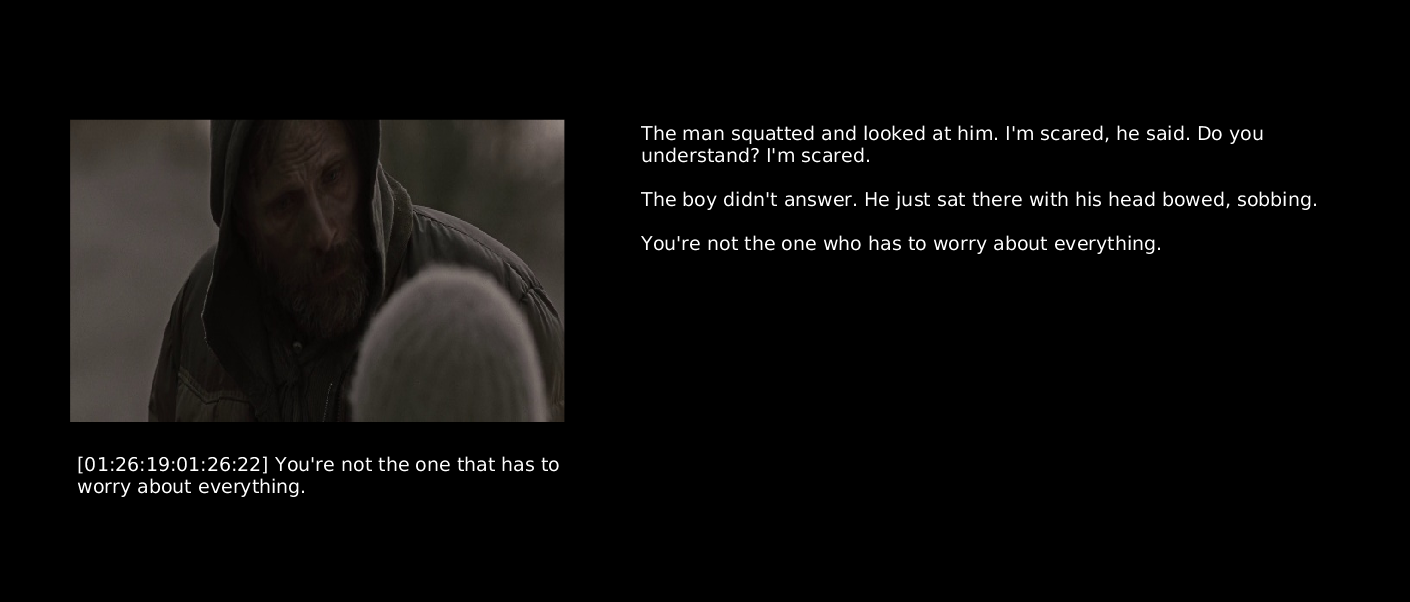}
\vspace{-2mm}
\caption{Describing movie clips via the book: we align the movie to the book, and show a shot from the movie and its corresponding paragraph (plus one before and after) from the book.}
\label{fig:moviecaption}
\vspace{1mm}
\end{figure*}

\begin{figure*}[t!]
\vspace{-3mm}
\centering
\hspace{-0.1mm}\includegraphics[height=0.192\linewidth,trim=40 25 55 25,clip]{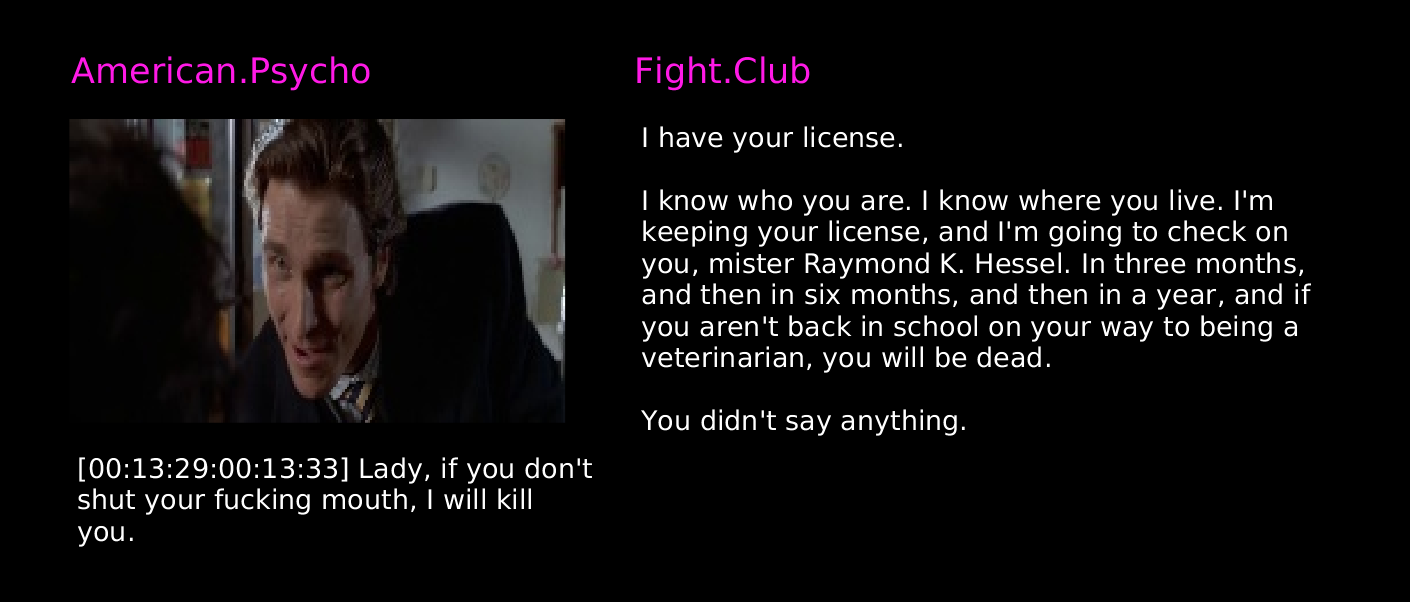}
\includegraphics[height=0.192\linewidth,trim=40 83 60 25,clip]{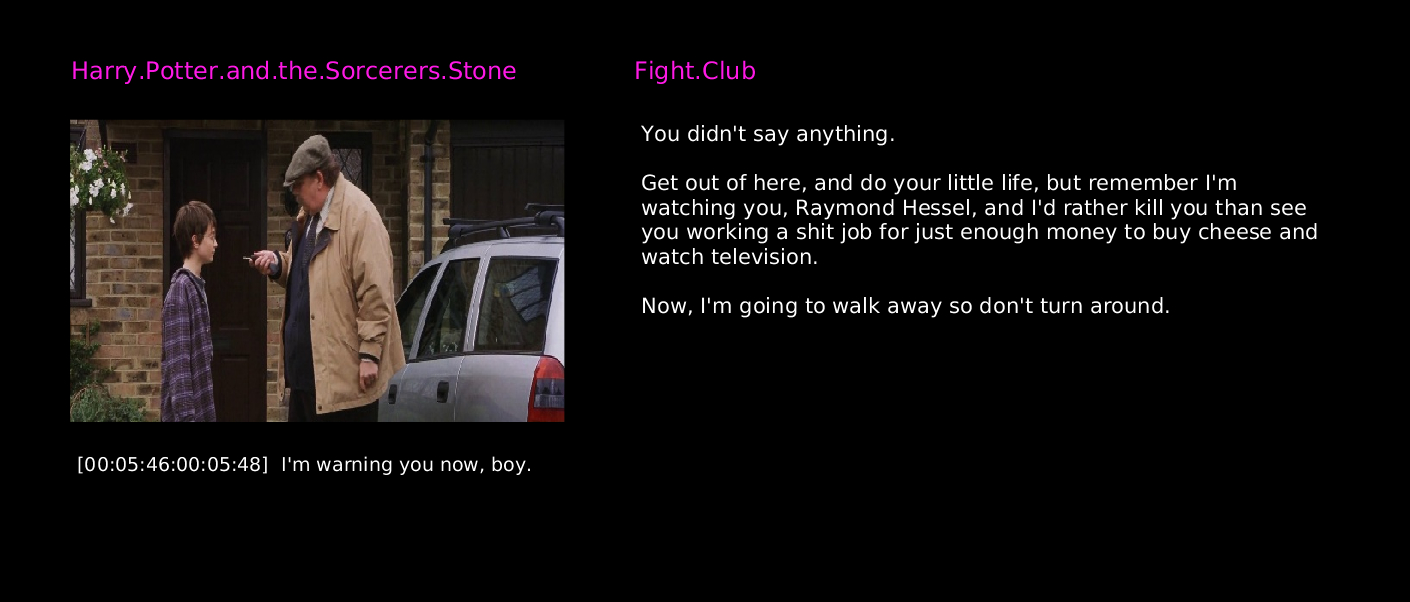}\\[0.5mm]
\hspace{-0.1mm}\includegraphics[height=0.196\linewidth,trim=38 55 55 25,clip]{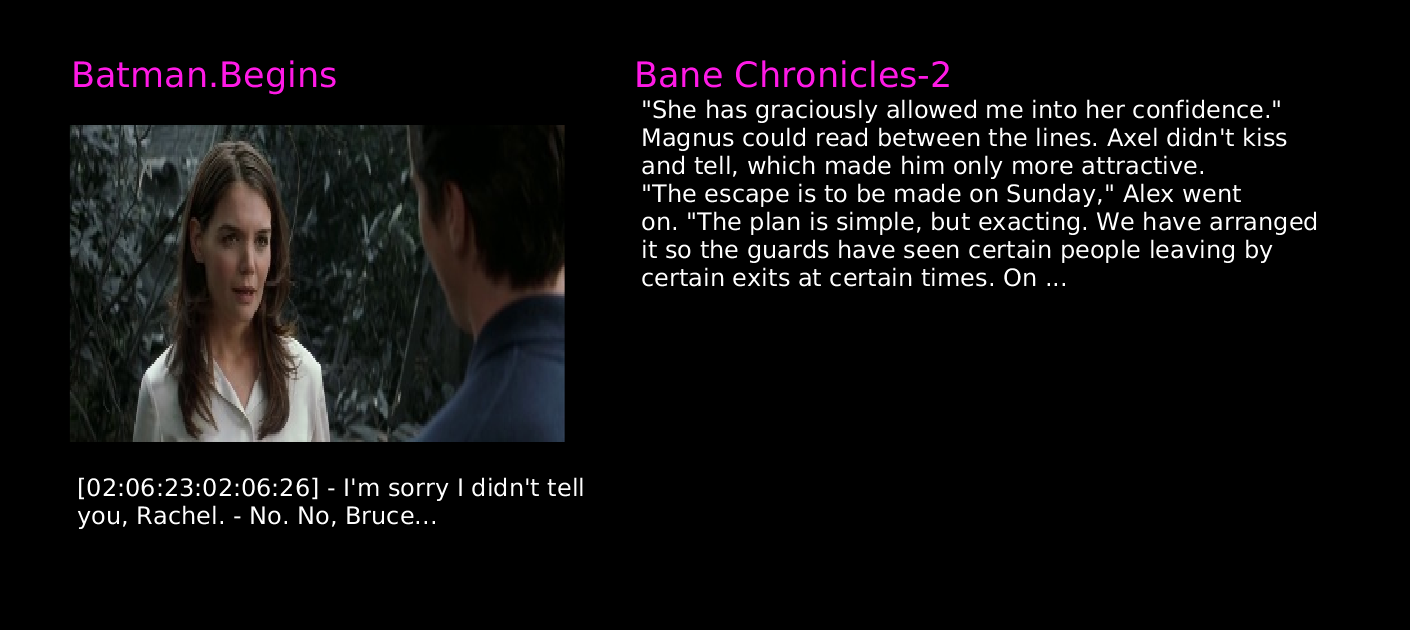}
\hspace{-0.1mm}\includegraphics[height=0.196\linewidth,trim=38 59 55 25,clip]{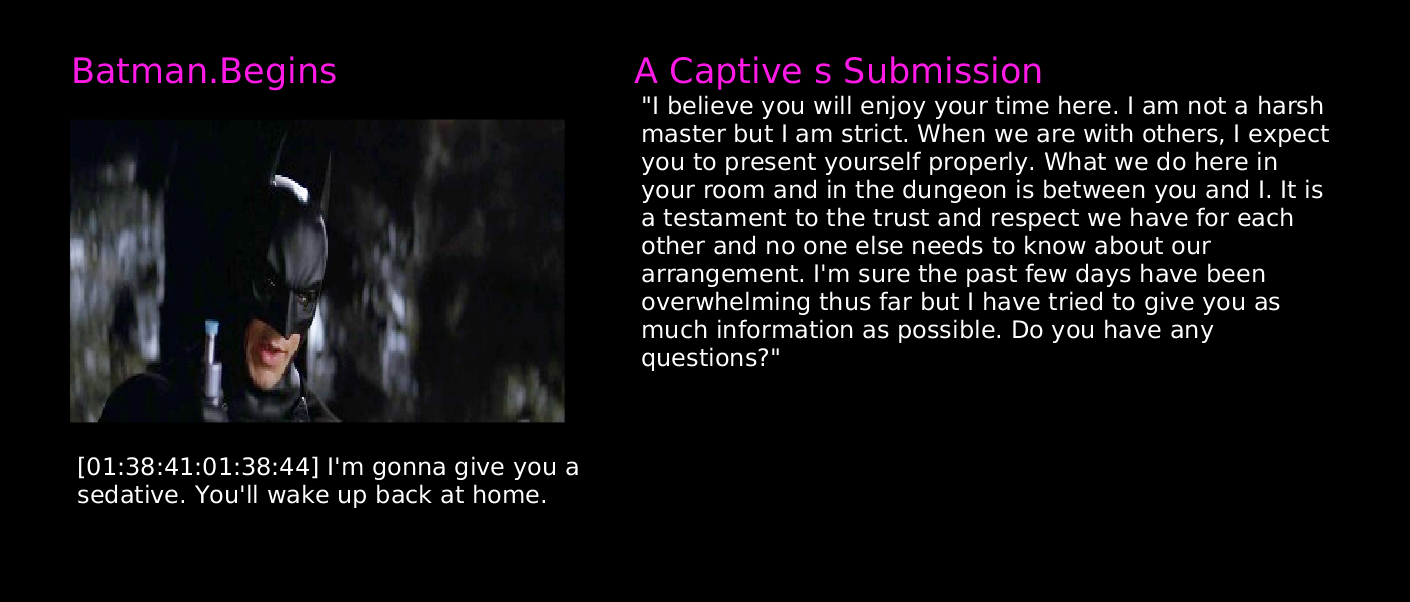}\\[0.5mm]
\hspace{-0.1mm}\includegraphics[height=0.190\linewidth,trim=38 55 55 25,clip]{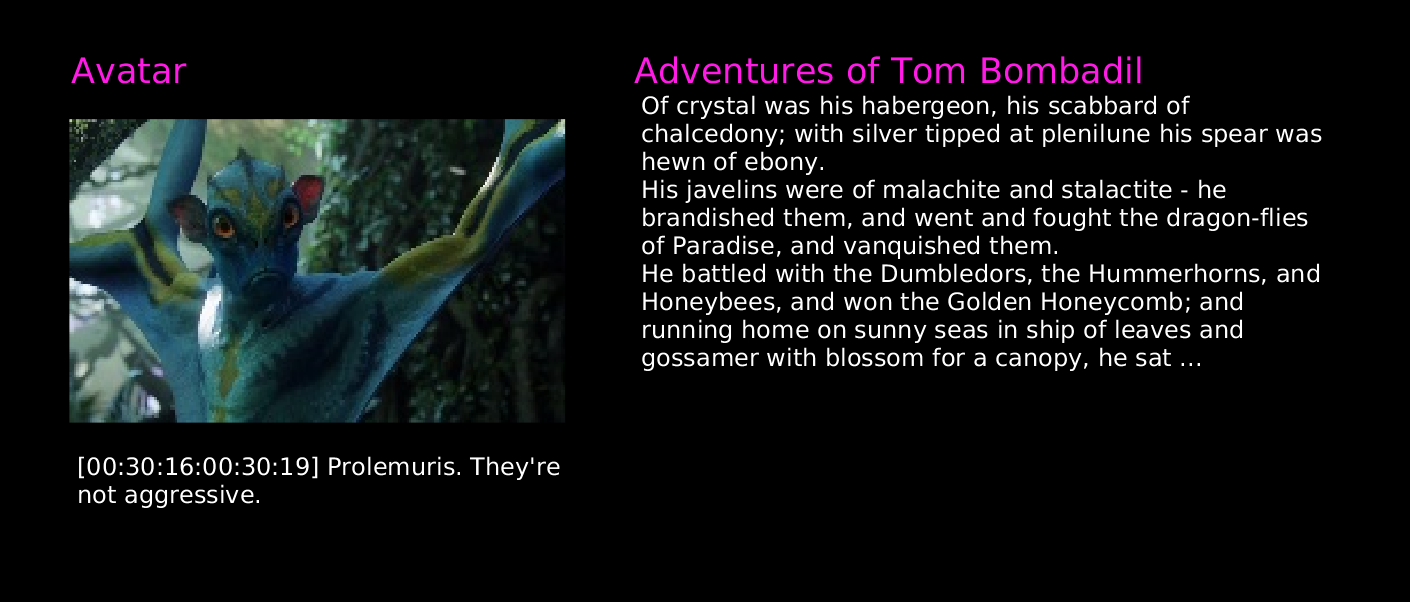}
\hspace{-0.1mm}\includegraphics[height=0.190\linewidth,trim=38 59 55 25,clip]{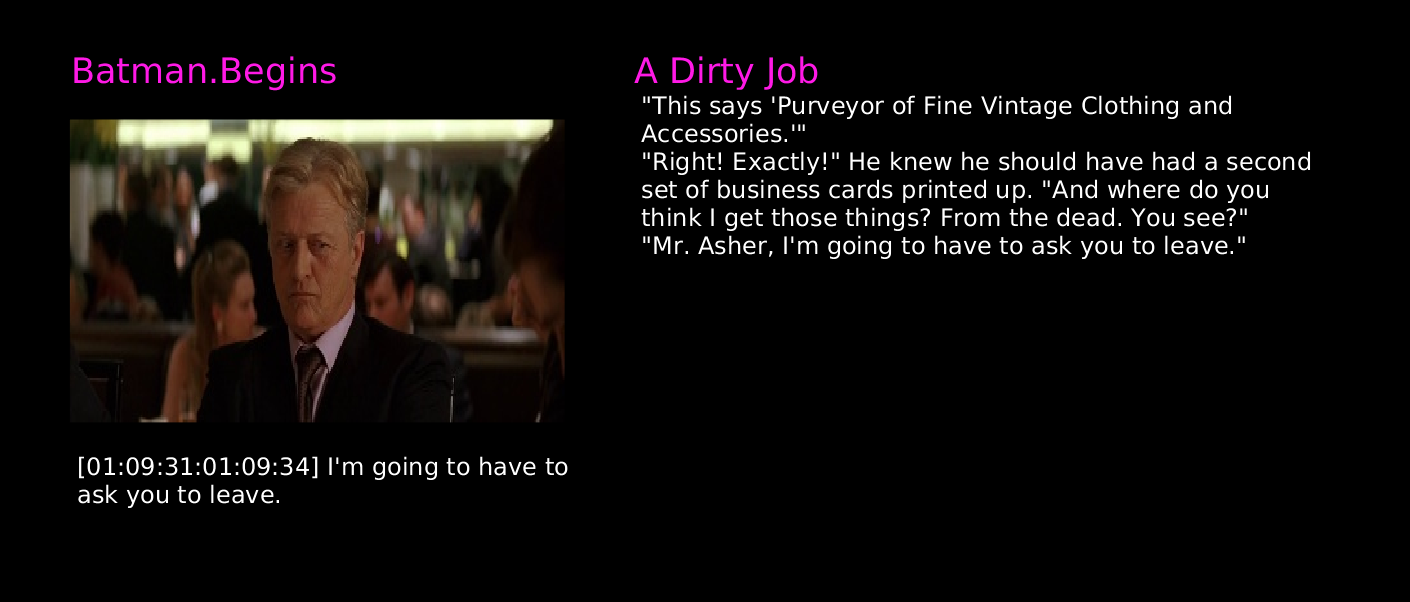}
\vspace{-2mm}
\caption{We can use our model to caption movies via a corpus of books. {\bf Top:} A shot from \emph{American Pyscho} is captioned with paragraphs from the \emph{Fight Club}, and a shot from \emph{Harry Potter} with paragraphs from \emph{Fight Club}. {\bf Middle} and {\bf Bottom:} We match shots from \emph{Avatar} and \emph{Batman Begins} against 300 books from our BookCorpus, and show the best matched paragraph.}
\label{fig:cross}
\vspace{-0mm}
\end{figure*}

\vspace{-1mm}
\subsection{Movie/Book Alignment}
\vspace{-1mm}

Evaluating the performance of movie/book alignment is an interesting problem on its own. 
This is because our ground-truth is far from exhaustive --  around 200 correspondences were typically found between a movie and its book, and likely a number of them got missed. Thus, evaluating the precision is rather tricky. We thus focus our evaluation on recall, similar to existing work on retrieval. For each shot that has a GT correspondence in book, we check whether our prediction is close to the annotated one. We evaluate recall at the paragraph level, i.e., we say that the GT paragraph was recalled, if our match was at most 3 paragraphs away, and the shot was at most 5 subtitle sentences away. As a noisier measure, we also compute recall and precision at multiple alignment thresholds and report AP (avg. prec.). 

The results are presented in Table~\ref{tb:comp}. Columns show different instantiations of our model: we show the leave-one-feature-out setting ($\emptyset$ indicates that all features were used), compare how different depths of the context-aware CNN influence the performance, and  compare it to our full model (CRF) in the last column. We get the highest boost by adding more layers to the CNN -- recall improves by $14\%$, and AP doubles. Generally, each feature helps  performance. Our sentence embedding (BOOK) helps by $4\%$, while noisier video-text embedding helps by $2\%$ in recall. CRF which encourages temporal smoothness generally helps (but not for all movies), bringing additional $2\%$. We also show how a  uniform timeline performs on its own.  That is, for each shot (measured in seconds) in the movie, we find the sentence at the same location (measured in lines) in the book. 

We add another baseline to evaluate the role of context in our model. Instead of using our CNN that considers contextual information, we build a linear SVM to combine different similarity measures in a single node (shot) -- the final similarity is used as a unary potential in our CRF alignment model. The Table shows that our CNN contextual model outperforms the SVM baseline by $30\%$ in recall, and doubles the AP.
 We plot alignment for a few movies in Fig.~\ref{fig:align}. 

\begin{table*}[htb!]\small
\footnotesize
\vspace{-1.5mm}
\addtolength{\tabcolsep}{0.75pt}
\begin{tabular}{|l|l|c|c|ccccccc|cc|}
\hline
 & & \multirow{2}{*}{UNI} & \multirow{2}{*}{SVM} & \multicolumn{7}{|c|}{$1$ layer CNN w/o one feature} & \multirow{2}{*}{CNN-$3$} & \multirow{2}{*}{CRF}  \\
\cline{5-11}
                   &   &  &            & $\emptyset$    & BLEU & TF-IDF  & BOOK & VIS & SCENE & PRIOR &  &         \\
                      \hline
\multirow{2}{0.95in}{Fight Club}                           & AP          & 1.22 & 0.73  & 0.45  & 0.41  & 0.40  & 0.50  & 0.64  & 0.50  & 0.48  & 1.95  & \textbf{5.17}   \\
                                     & Recall      & 2.36  & 10.38 & 12.26 & 12.74 & 11.79 & 11.79 & 12.74 & 11.79 & 11.79 & 17.92 & \textbf{19.81}  \\
                                       \hline
\multirow{2}{0.95in}{The Green Mile }                      & AP          & 0.00   & 14.05 & 14.12 & 14.09 & 6.92  & 10.12 & 9.83  & 13.00 & 14.42 & \textbf{28.80} & 27.60  \\
                                     & Recall      & 0.00  & 51.42 & 62.46 & 60.57 & 53.94 & 57.10 & 55.52 & 60.57 & 62.78 & 74.13 & \textbf{78.23}  \\
                                       \hline
\multirow{2}{0.95in}{Harry Potter and the Sorcerers Stone }& AP          & 0.00  &10.30 & 8.09  & 8.18  & 5.66  & 7.84  & 7.95  & 8.04  & 8.20  & \textbf{27.17} & 23.65  \\
                                     & Recall      & 0.00  &44.35 & 51.05 & 52.30 & 46.03 & 48.54 & 48.54 & 49.37 & 52.72 & 76.57 & \textbf{78.66}  \\
                                       \hline
\multirow{2}{0.95in}{American Psycho    }                  & AP          & 0.00   & 14.78 & 16.76 & 17.22 & 12.29 & 14.88 & 14.95 & 15.68 & 16.54 & \textbf{34.32} & 32.87   \\
                                     & Recall      & 0.27  & 34.25 & 67.12 & 66.58 & 60.82 & 64.66 & 63.56 & 66.58 & 67.67 & \textbf{81.92} & 80.27  \\
                                       \hline
\multirow{2}{0.95in}{One Flew Over the Cuckoo Nest  }      & AP          & 0.00   & 5.68  & 8.14  & 6.27  & 1.93  & 8.49  & 8.51  & 9.32  & 9.04  & 14.83 & \textbf{21.13}  \\
                                     & Recall      & 1.01  &25.25 & 41.41 & 34.34 & 32.32 & 36.36 & 37.37 & 36.36 & 40.40 & 49.49 & \textbf{54.55}  \\
                                       \hline
\multirow{2}{0.95in}{Shawshank Redemption  }               & AP          & 0.00   & 8.94  & 8.60  & 8.89  & 4.35  & 7.99  & 8.91  & 9.22  & 7.86  & 19.33 & \textbf{19.96}  \\
                                     & Recall    &  1.79 & 46.43 & 78.57 & 76.79 & 73.21 & 73.21 & 78.57 & 75.00 & 78.57 & 94.64 & \textbf{96.79}  \\
                                       \hline
\multirow{2}{0.95in}{The Firm     }                        & AP          & 0.05   & 4.46  & 7.91  & 8.66  & 2.02  & 6.22  & 7.15  & 7.25  & 7.26  & 18.34 & \textbf{20.74}  \\
                                     & Recall      & 1.38  &18.62 & 33.79 & 36.55 & 26.90 & 23.45 & 26.90 & 30.34 & 31.03 & 37.93 & \textbf{44.83}  \\
                                       \hline
\multirow{2}{0.95in}{Brokeback Mountain     }              & AP          & 2.36   & 24.91 & 16.55 & 17.82 & 14.60 & 15.16 & 15.58 & 15.41 & 16.21 & \textbf{31.80} & 30.58  \\
                                     & Recall   &  27.0 & 74.00 & 88.00 & 92.00 & 86.00 & 86.00 & 88.00 & 86.00 & 87.00 & 98.00 & \textbf{100.00} \\
                                       \hline
\multirow{2}{0.95in}{The Road  }                           & AP          & 0.00   & 13.77 & 6.58  & 7.83  & 3.04  & 5.11  & 5.47  & 6.09  & 7.00  & \textbf{19.80} & 19.58  \\
                                     & Recall   & 1.12   & 41.90 & 43.02 & 48.04 & 32.96 & 38.55 & 37.99 & 42.46 & 44.13 & \textbf{65.36} & 65.10  \\
                                     \hline
\multirow{2}{0.95in}{No Country for Old Men }                           & AP          & 0.00   & 12.11 & 9.00  & 9.39  & 8.22  & 9.40  & 9.35  & 8.63  & 9.40  & 28.75 & \textbf{30.45}  \\
                                     & Recall      & 1.12  & 33.46 & 48.90 & 49.63 & 46.69 & 47.79 & 51.10 & 49.26 & 48.53 & 71.69 & \textbf{72.79}  \\
                                     \hline
                                     \hline
\multicolumn{2}{|c|}{Mean Recall}  & 3.88& 38.01 & 52.66 & 52.95 & 47.07 & 48.75 & 50.03 & 50.77 & 52.46 & 66.77 & \textbf{69.10} \\

\multicolumn{2}{|c|}{AP}   &  0.40  & 10.97 & 9.62  & 9.88  & 5.94  & 8.57  & 8.83  & 9.31  & 9.64  & 22.51 & \textbf{23.17} \\
\hline
\end{tabular}
\vspace{-3mm}
\caption{Performance of our model for the movies in our dataset under different settings and metrics.}\label{tb:comp}
\vspace{-0.5mm}
\end{table*}

\begin{table*}[t!]
\footnotesize
\begin{tabular}{|l|c|c|c|c|c|c|c|c|c|}
\hline
      & BLEU    & TF     & BOOK  & VIS     & SCENE  & CNN (training) & CNN (inference) & CRF (training) & CRF (inference) \\ \hline
Per movie-book pair & 6h & 10 min & 3 min & 2h & 1h & 3 min      & 0.2 min   & 5h    & 5 min   \\
\hline
\end{tabular}
\vspace{-3mm}
\caption{Running time for our model per one movie/book pair.}\label{tb:time}
\vspace{-4mm}
\end{table*}

\begin{table}[t!]\tiny
\vspace{-2mm}
\centering
\hspace{-2mm}\addtolength{\tabcolsep}{-4.8pt}
\begin{tabular}{|c|p{1.1mm}cp{1.1mm}cp{1.1mm}cp{1.1mm}cp{1.1mm}cp{1.1mm}cp{1.1mm}c|}
\hline
\multicolumn{1}{|>{\columncolor{bgr}}c|}{} & \multicolumn{14}{>{\columncolor{gr}}c|}{}\\[-0.6mm]
\multicolumn{1}{|>{\columncolor{bgr}}c|}{\sc\small movie} & \multicolumn{14}{>{\columncolor{gr}}c|}{\sc\small books}\\[-0.6mm]
\multicolumn{1}{|>{\columncolor{bgr}}c|}{\tiny} & \multicolumn{14}{>{\columncolor{gr}}c|}{\tiny}\\
\hline
\includegraphics[width=0.044\textwidth, height=9.4mm]{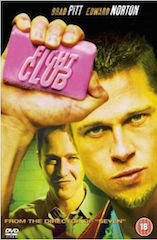}
 & \rotatebox{90}{Fight Club} & \includegraphics[width=0.044\textwidth, height=9.4mm]{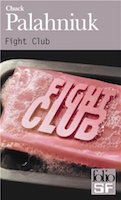} 
 & \rotatebox{90}{No Country ...} & \includegraphics[width=0.044\textwidth, height=9.4mm]{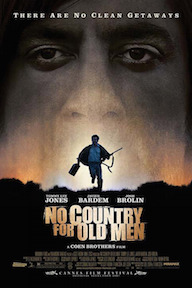} 
 & \rotatebox{90}{One Flew...} & \includegraphics[width=0.044\textwidth, height=9.4mm]{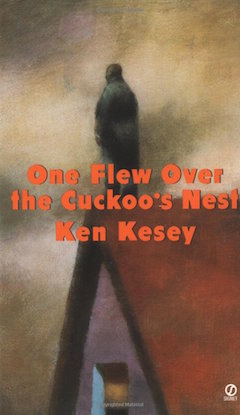} 
 & \rotatebox{90}{The Road} & \includegraphics[width=0.044\textwidth, height=9.4mm]{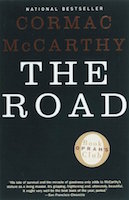} 
 & \rotatebox{90}{The Firm} & \includegraphics[width=0.044\textwidth, height=9.4mm]{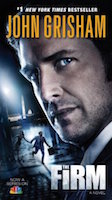} 
 & \rotatebox{90}{American Psy.} & \includegraphics[width=0.044\textwidth, height=9.4mm]{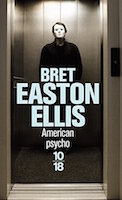} 
 & \rotatebox{90}{Shawshank ...} & \includegraphics[width=0.044\textwidth, height=9.4mm]{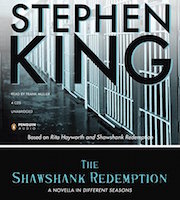} \\[-0.5mm]
Fight Club    & & 100.0 & & 45.4 & & 45.2 & & 45.1 & & 43.6 & & 43.0 & & 42.7\\[-0.5mm]
 \includegraphics[width=0.044\textwidth, height=9.4mm]{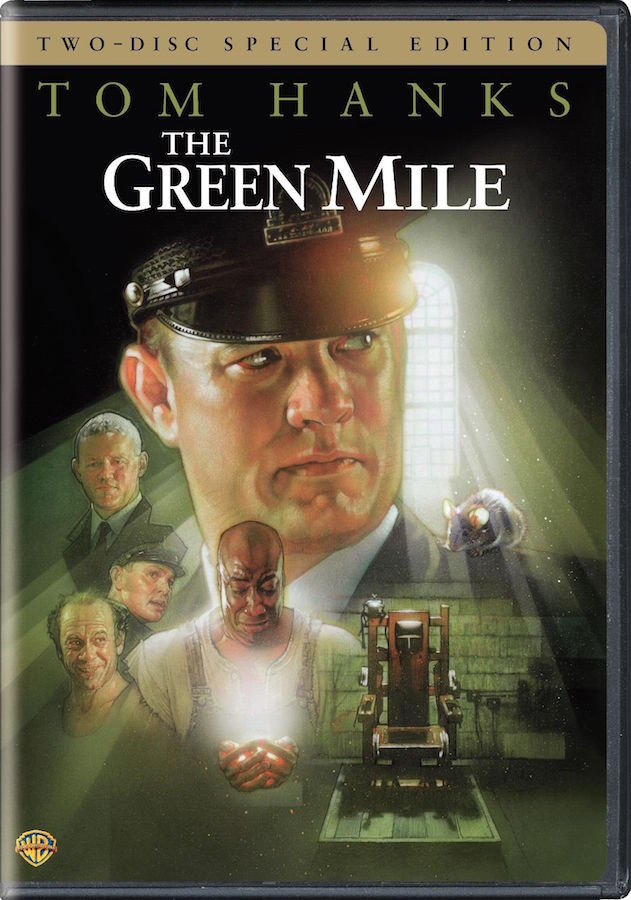}
 & \rotatebox{90}{Green Mile} & \includegraphics[width=0.044\textwidth, height=9.4mm]{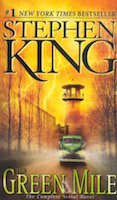} 
 & \rotatebox{90}{One Flew...} & \includegraphics[width=0.044\textwidth, height=9.4mm]{figs/covers/cukoo_book1.jpg} 
 & \rotatebox{90}{American Psy.} & \includegraphics[width=0.044\textwidth, height=9.4mm]{figs/covers/americanpsycho_book.jpg} 
 & \rotatebox{90}{Shawshank ...} & \includegraphics[width=0.044\textwidth, height=9.4mm]{figs/covers/shawshank_book.jpg} 
 & \rotatebox{90}{No Country ...} & \includegraphics[width=0.044\textwidth, height=9.4mm]{figs/covers/nocountry_dvd.jpg} 
 & \rotatebox{90}{The Firm} & \includegraphics[width=0.044\textwidth, height=9.4mm]{figs/covers/thefirm_book.jpg} 
 & \rotatebox{90}{The Road} & \includegraphics[width=0.044\textwidth, height=9.4mm]{figs/covers/theroad_book.jpg} \\[-0.5mm]
 Green Mile  & & 100.0 & & 42.5 & & 40.1 & & 39.6 & & 38.9 & & 38.0 & & 36.7\\[-0.5mm]
  \includegraphics[width=0.044\textwidth, height=9.4mm]{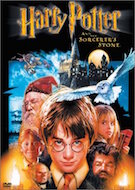}
& \rotatebox{90}{Harry Potter} & \includegraphics[width=0.044\textwidth, height=9.4mm]{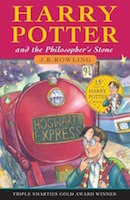} 
 & \rotatebox{90}{Brokeback ...} & \includegraphics[width=0.044\textwidth, height=9.4mm]{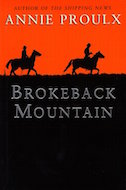} 
 & \rotatebox{90}{American Psy.} & \includegraphics[width=0.044\textwidth, height=9.4mm]{figs/covers/americanpsycho_book.jpg} 
 & \rotatebox{90}{The Road} & \includegraphics[width=0.044\textwidth, height=9.4mm]{figs/covers/theroad_book.jpg} 
 & \rotatebox{90}{One Flew...} & \includegraphics[width=0.044\textwidth, height=9.4mm]{figs/covers/cukoo_book1.jpg} 
 & \rotatebox{90}{Shawshank ...} & \includegraphics[width=0.044\textwidth, height=9.4mm]{figs/covers/shawshank_book.jpg} 
 & \rotatebox{90}{The Firm} & \includegraphics[width=0.044\textwidth, height=9.4mm]{figs/covers/thefirm_book.jpg} \\[-0.5mm]
 Harry Potter  & & 100.0 & & 40.5 & & 39.7 & & 39.5 & & 39.1 & & 39.0 & & 38.7\\[-1.2mm]
  \includegraphics[width=0.044\textwidth, height=9.4mm]{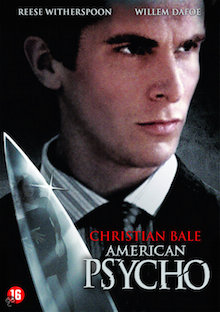} 
 & \rotatebox{90}{American Psy.o} & \includegraphics[width=0.044\textwidth, height=9.4mm]{figs/covers/americanpsycho_book.jpg} 
 & \rotatebox{90}{The Firm} & \includegraphics[width=0.044\textwidth, height=9.4mm]{figs/covers/thefirm_book.jpg} 
 & \rotatebox{90}{One Flew...} & \includegraphics[width=0.044\textwidth, height=9.4mm]{figs/covers/cukoo_book1.jpg} 
 & \rotatebox{90}{Fight Club} & \includegraphics[width=0.044\textwidth, height=9.4mm]{figs/covers/fightclub_book.jpg} 
 & \rotatebox{90}{Shawshank ...} & \includegraphics[width=0.044\textwidth, height=9.4mm]{figs/covers/shawshank_book.jpg} 
 & \rotatebox{90}{No Country ...} & \includegraphics[width=0.044\textwidth, height=9.4mm]{figs/covers/nocountry_dvd.jpg} 
 & \rotatebox{90}{Brokeback ...} & \includegraphics[width=0.044\textwidth, height=9.4mm]{figs/covers/brokebackmountain_book.jpg}\\[-0.5mm]
 American Psy.  & & 100.0 & & 55.5 & & 54.9 & & 53.5 & & 53.1 & & 52.6 & & 51.3\\[-0.5mm]
   \includegraphics[width=0.044\textwidth, height=9.4mm]{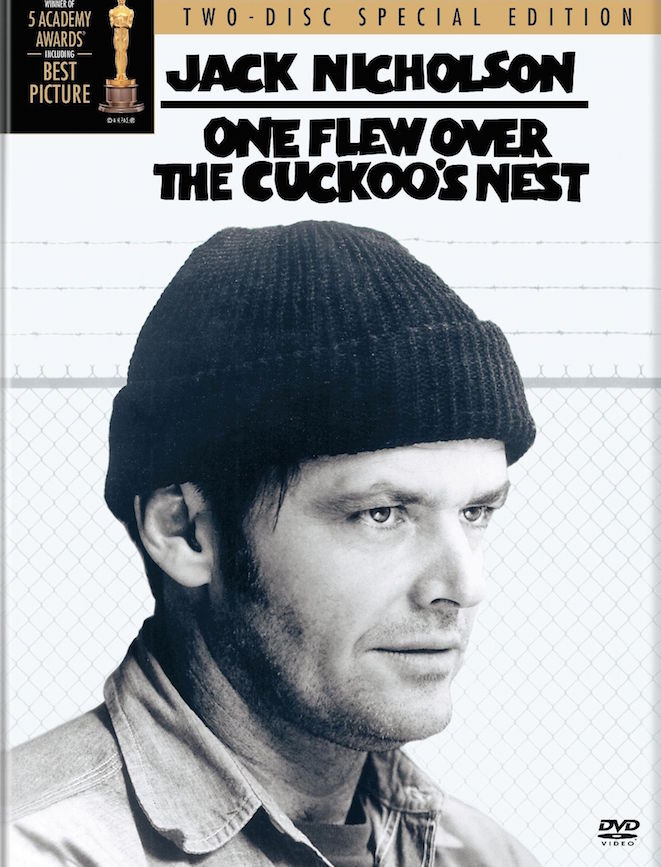} 
 & \rotatebox{90}{One Flew...} & \includegraphics[width=0.044\textwidth, height=9.4mm]{figs/covers/cukoo_book1.jpg} 
 & \rotatebox{90}{The Firm} & \includegraphics[width=0.044\textwidth, height=9.4mm]{figs/covers/thefirm_book.jpg} 
 & \rotatebox{90}{Harry Potter} & \includegraphics[width=0.044\textwidth, height=9.4mm]{figs/covers/harrypotter_book.jpg} 
 & \rotatebox{90}{The Road} & \includegraphics[width=0.044\textwidth, height=9.4mm]{figs/covers/theroad_book.jpg} 
 & \rotatebox{90}{Shawshank ...} & \includegraphics[width=0.044\textwidth, height=9.4mm]{figs/covers/shawshank_book.jpg} 
 & \rotatebox{90}{Brokeback ...} & \includegraphics[width=0.044\textwidth, height=9.4mm]{figs/covers/brokebackmountain_book.jpg} 
 & \rotatebox{90}{No Country ...} & \includegraphics[width=0.044\textwidth, height=9.4mm]{figs/covers/nocountry_dvd.jpg} \\[-0.5mm]
 One Flew...  & & 100.0 & & 84.0 & & 80.8 & & 79.1 & & 79.0 & & 77.8 & & 76.9\\[-0.5mm]
    \includegraphics[width=0.044\textwidth, height=9.4mm]{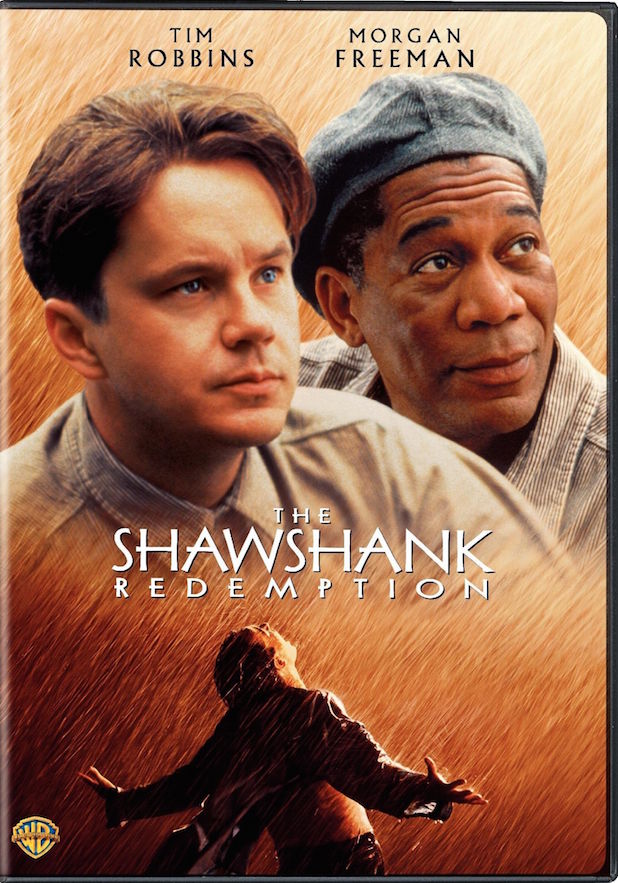}  
 & \rotatebox{90}{Shawshank ...} & \includegraphics[width=0.044\textwidth, height=9.4mm]{figs/covers/shawshank_book.jpg} 
 & \rotatebox{90}{The Firm} & \includegraphics[width=0.044\textwidth, height=9.4mm]{figs/covers/thefirm_book.jpg} 
 & \rotatebox{90}{No Country ...} & \includegraphics[width=0.044\textwidth, height=9.4mm]{figs/covers/nocountry_dvd.jpg} 
 & \rotatebox{90}{One Flew...} & \includegraphics[width=0.044\textwidth, height=9.4mm]{figs/covers/cukoo_book1.jpg} 
 & \rotatebox{90}{The Road} & \includegraphics[width=0.044\textwidth, height=9.4mm]{figs/covers/theroad_book.jpg} 
 & \rotatebox{90}{Brokeback ...} & \includegraphics[width=0.044\textwidth, height=9.4mm]{figs/covers/brokebackmountain_book.jpg} 
 & \rotatebox{90}{Green Mile} & \includegraphics[width=0.044\textwidth, height=9.4mm]{figs/covers/greenmile_book.jpg} \\[-0.5mm]
 Shawshank ...  & & 100.0 & & 66.0 & & 62.0 & & 61.4 & & 60.9 & & 59.1 & & 58.0\\[-0.5mm]
  \includegraphics[width=0.044\textwidth, height=9.4mm]{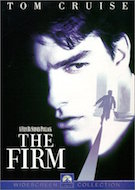}   
 & \rotatebox{90}{The Firm} & \includegraphics[width=0.044\textwidth, height=9.4mm]{figs/covers/thefirm_book.jpg} 
  & \rotatebox{90}{Shawshank ...} & \includegraphics[width=0.044\textwidth, height=9.4mm]{figs/covers/shawshank_book.jpg} 
  & \rotatebox{90}{Fight Club} & \includegraphics[width=0.044\textwidth, height=9.4mm]{figs/covers/fightclub_book.jpg} 
  & \rotatebox{90}{Brokeback ...} & \includegraphics[width=0.044\textwidth, height=9.4mm]{figs/covers/brokebackmountain_book.jpg} 
  & \rotatebox{90}{One Flew...} & \includegraphics[width=0.044\textwidth, height=9.4mm]{figs/covers/cukoo_book1.jpg} 
  & \rotatebox{90}{American Psy.} & \includegraphics[width=0.044\textwidth, height=9.4mm]{figs/covers/americanpsycho_book.jpg} 
  & \rotatebox{90}{Harry Potter} & \includegraphics[width=0.044\textwidth, height=9.4mm]{figs/covers/harrypotter_book.jpg} \\[-0.5mm]
  The Firm  & & 100.0 & & 75.0 & & 73.9 & & 73.7 & & 71.5 & & 71.4 & & 68.5\\[-0.5mm]
      \includegraphics[width=0.044\textwidth, height=9.4mm]{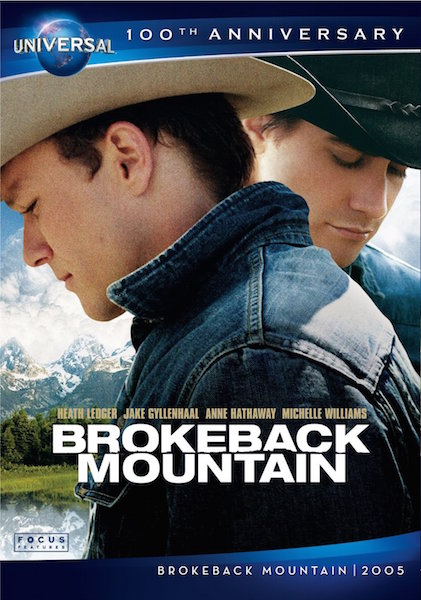}  
  & \rotatebox{90}{Brokeback ...} & \includegraphics[width=0.044\textwidth, height=9.4mm]{figs/covers/brokebackmountain_book.jpg} 
 & \rotatebox{90}{One Flew...} & \includegraphics[width=0.044\textwidth, height=9.4mm]{figs/covers/cukoo_book1.jpg} 
 & \rotatebox{90}{Fight Club} & \includegraphics[width=0.044\textwidth, height=9.4mm]{figs/covers/fightclub_book.jpg} 
 & \rotatebox{90}{American Psy.} & \includegraphics[width=0.044\textwidth, height=9.4mm]{figs/covers/americanpsycho_book.jpg} 
 & \rotatebox{90}{Green Mile} & \includegraphics[width=0.044\textwidth, height=9.4mm]{figs/covers/greenmile_book.jpg} 
 & \rotatebox{90}{The Firm} & \includegraphics[width=0.044\textwidth, height=9.4mm]{figs/covers/thefirm_book.jpg} 
 & \rotatebox{90}{Shawshank ...} & \includegraphics[width=0.044\textwidth, height=9.4mm]{figs/covers/shawshank_book.jpg} \\[-0.5mm]
 Brokeback ...  & & 100.0 & & 54.8 & & 52.2 & & 51.9 & & 50.9 & & 50.7 & & 50.6\\[-0.5mm]
      \includegraphics[width=0.044\textwidth, height=9.4mm,trim=50 0 50 0,clip]{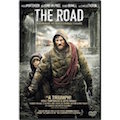}  
 & \rotatebox{90}{The Road} & \includegraphics[width=0.044\textwidth, height=9.4mm]{figs/covers/theroad_book.jpg} 
 & \rotatebox{90}{The Firm} & \includegraphics[width=0.044\textwidth, height=9.4mm]{figs/covers/thefirm_book.jpg} 
 & \rotatebox{90}{One Flew...} & \includegraphics[width=0.044\textwidth, height=9.4mm]{figs/covers/cukoo_book1.jpg} 
 & \rotatebox{90}{No Country ...} & \includegraphics[width=0.044\textwidth, height=9.4mm]{figs/covers/nocountry_dvd.jpg} 
 & \rotatebox{90}{Fight Club} & \includegraphics[width=0.044\textwidth, height=9.4mm]{figs/covers/fightclub_book.jpg} 
 & \rotatebox{90}{Shawshank ...} & \includegraphics[width=0.044\textwidth, height=9.4mm]{figs/covers/shawshank_book.jpg} 
 & \rotatebox{90}{Green Mile} & \includegraphics[width=0.044\textwidth, height=9.4mm]{figs/covers/greenmile_book.jpg} \\[-0.5mm]
 The Road  & & 100.0 & & 56.0 & & 55.9 & & 54.8 & & 54.1 & & 53.9 & & 53.4\\[-0.5mm]
      \includegraphics[width=0.044\textwidth, height=9.4mm]{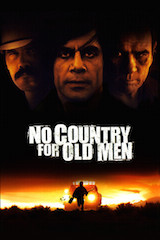}  
& \rotatebox{90}{No Country ...} & \includegraphics[width=0.044\textwidth, height=9.4mm]{figs/covers/nocountry_dvd.jpg} 
 & \rotatebox{90}{The Road} & \includegraphics[width=0.044\textwidth, height=9.4mm]{figs/covers/theroad_book.jpg} 
 & \rotatebox{90}{Brokeback ...} & \includegraphics[width=0.044\textwidth, height=9.4mm]{figs/covers/brokebackmountain_book.jpg} 
 & \rotatebox{90}{One Flew...} & \includegraphics[width=0.044\textwidth, height=9.4mm]{figs/covers/cukoo_book1.jpg} 
 & \rotatebox{90}{The Firm} & \includegraphics[width=0.044\textwidth, height=9.4mm]{figs/covers/thefirm_book.jpg} 
 & \rotatebox{90}{Shawshank ...} & \includegraphics[width=0.044\textwidth, height=9.4mm]{figs/covers/shawshank_book.jpg} 
 & \rotatebox{90}{Harry Potter} & \includegraphics[width=0.044\textwidth, height=9.4mm]{figs/covers/harrypotter_book.jpg} \\[-0.5mm]
 No Country...  & & 100.0 & & 49.7 & & 49.5 & & 46.8 & & 46.4 & & 45.8 & & 45.8\\

\hline
\end{tabular}
\vspace{-2mm}
\caption{{\bf Book ``retrieval''}. For a movie (left), we rank  books wrt to their alignment similarity with the movie.  We normalize similarity to be 100 for the highest scoring book.}
\label{fig:retrieval}
\vspace{-2mm}
\end{table}

\vspace{-3mm}
\paragraph{Running Times.}
We show the typical running time of each component in our model in Table~\ref{tb:time}. For each movie-book pair, calculating BLEU score takes most of the time. Note that BLEU does not contribute significantly to the performance and is of optional use. With respect to the rest,  extracting visual features VIS (mean pooling GoogleNet features over the shot frames) and SCENE features (mean pooling hybrid-CNN features~\cite{ZhouNIPS2014} over the shot frames), takes most of the time (about $80\%$ of the total time). 

We also report training times for our contextual model (CNN) and the CRF alignment model. Note that the times are reported for one movie/book pair since we used only one such pair to train  all our CNN and CRF parameters. We chose \emph{Gone Girl} for training since it had the best balance between the dialog and visual correspondences.

\vspace{-1mm}
\subsection{Describing Movies via the Book}
\vspace{-1mm}

We next show qualitative results of our alignment. In particular, we run our model on each movie/book pair, and visualize the passage in the book that a particular shot in the movie aligns to. We show  best matching paragraphs as well as a paragraph before and after. The results are shown in Fig.~\ref{fig:align}. One can see that our model is able to retrieve a semantically meaningful match despite large dialog deviations from those in the book, and the challenge of matching a visual representation to the verbose text in the book.

\renewcommand{\baselinestretch}{0.42}
\begin{figure*}
\vspace{-3.5mm}
\begin{minipage}{0.255\textwidth}
\includegraphics[height=2.96cm]{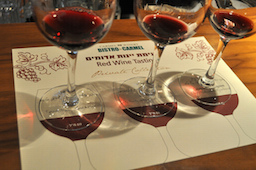}\\
\label{fig:cocobook}
\begin{tiny}
the club was a little emptier than i would have expected for the late afternoon , and the bartender , in red waistcoat and bowtie , was busy wiping down his counter , replacing peanuts and putting out new coasters . 
a television with the latest la liga news was hung in an upper corner , and behind him , rows of bottles were reflected in a giant bar mirror . 
above the stools , a pergola-type overhead structure held rows of wine glasses . 
it was a classy place , with ferns in the corner , and not the kind of bar to which i was accustomed . 
my places usually had a more ... relaxed feel .
\end{tiny}
\end{minipage} \hspace{0.2mm}
\begin{minipage}{0.218\textwidth}
\includegraphics[height=2.98cm]{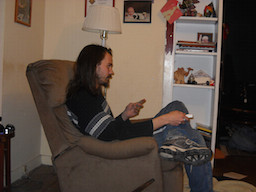}\\[0.0mm]
\begin{tiny}
he felt like an idiot for yelling at the child , but his frustration and trepidation was getting the better of him .
he glanced toward the shadowed hall and quickly nodded toward melissa before making his way forward .
he came across more children sitting upon a couch in the living room .
they watched him , but did n't move and did n't speak .
his skin started to feel like hundreds of tiny spiders were running up and down it and he hurried on .\\
$\ $
\end{tiny}
\end{minipage}\hspace{2.7mm}
\begin{minipage}{0.218\textwidth}
\includegraphics[height=2.96cm]{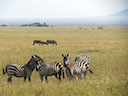}\\
\begin{tiny}
a few miles before tioga road reached highway 395 and the town of lee vining , smith turned onto a narrow blacktop road .
on either side were parched , grassy open slopes with barbed-wire fences marking property lines .
cattle and horses grazed under trees whose black silhouettes stood stark against the gold-velvet mountains .
marty burst into song : `` home , home on the range , where the deer and the antelope play !
where seldom is heard a discouraging word and the skies are not cloudy all day ! ''
\end{tiny}
\end{minipage}\hspace{1.5mm}
\renewcommand{\baselinestretch}{0.4}
\begin{minipage}{0.258\textwidth}
\includegraphics[width=0.93\textwidth]{figs/city}
\begin{tiny}
``number seventy-three , second to last from the corner . '
adam slowed the porsche as he approached the quaint-he could think of no other word to use , even though ``quaint'' was one he normally , manfully , avoided-townhouse , coming to a halt beside a sleek jaguar sedan .
it was a quiet street , devoid of traffic at this hour on a monday night .
in the bluish-tinted light of a corner street lamp , he developed a quick visual impression of wrought-iron railings on tidy front stoops , window boxes full of bright chrysanthemums , beveled glass in bay windows , and lace curtains .
townhouses around here didn't rent cheaply , he could n't help but observe .
\end{tiny}
\end{minipage}
\renewcommand{\baselinestretch}{1}
\vspace{-2mm}
\caption{CoCoBook: We generate a caption for a CoCo image via~\cite{DBLP:journals/corr/KirosSZ14} and retrieve its best matched sentence (+ 2 before and after) from a large book corpus. One can see a semantic relevance of the retrieved passage to the image.}
\label{fig:cocobook}
\end{figure*}
\renewcommand{\baselinestretch}{1}

\begin{figure*}[htb]
\begin{center}    
\begin{minipage}{1\linewidth}
                \includegraphics[width=0.332\textwidth]{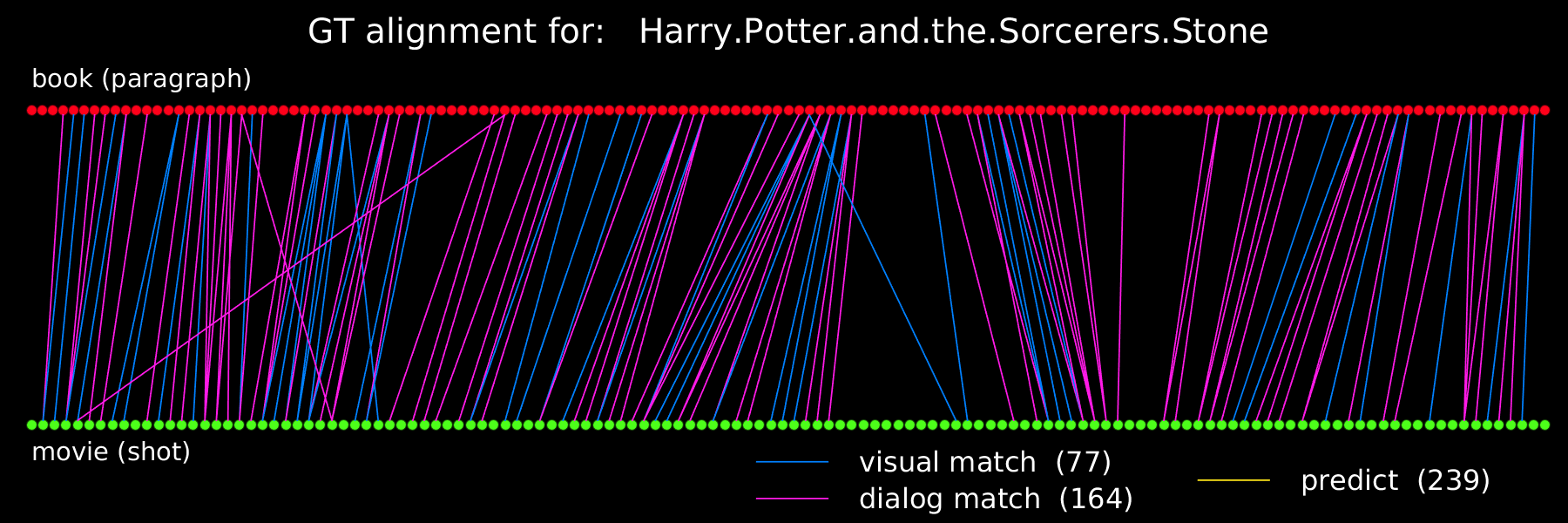}
                \includegraphics[width=0.332\textwidth]{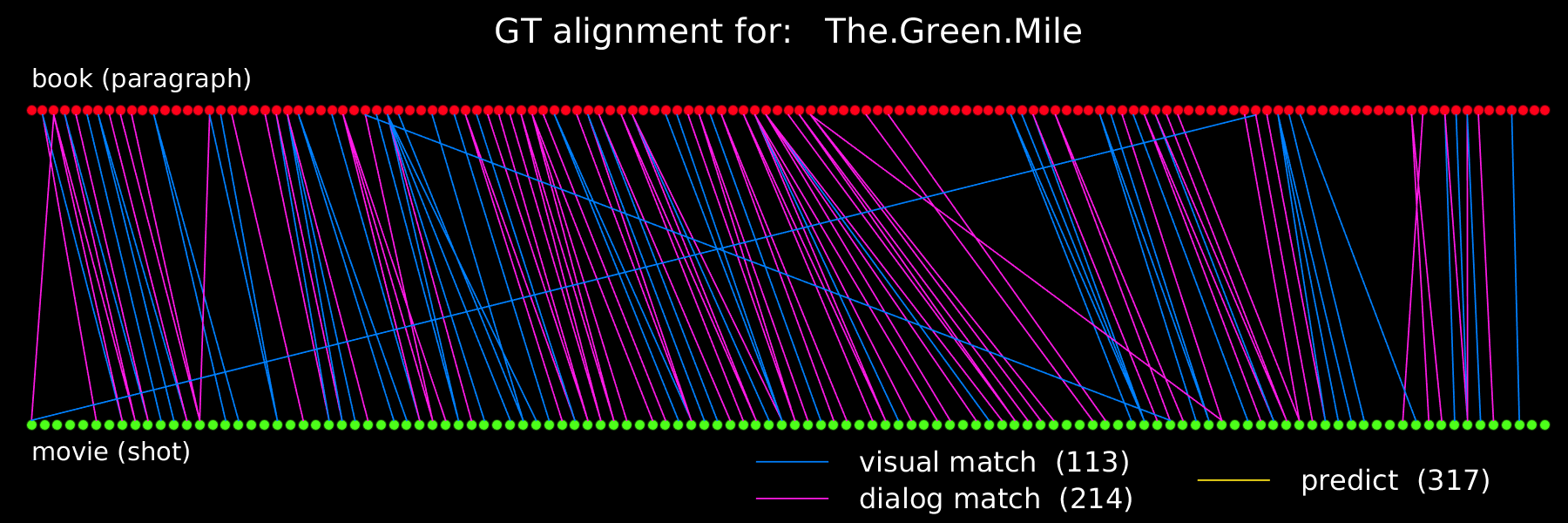}
                \includegraphics[width=0.332\textwidth]{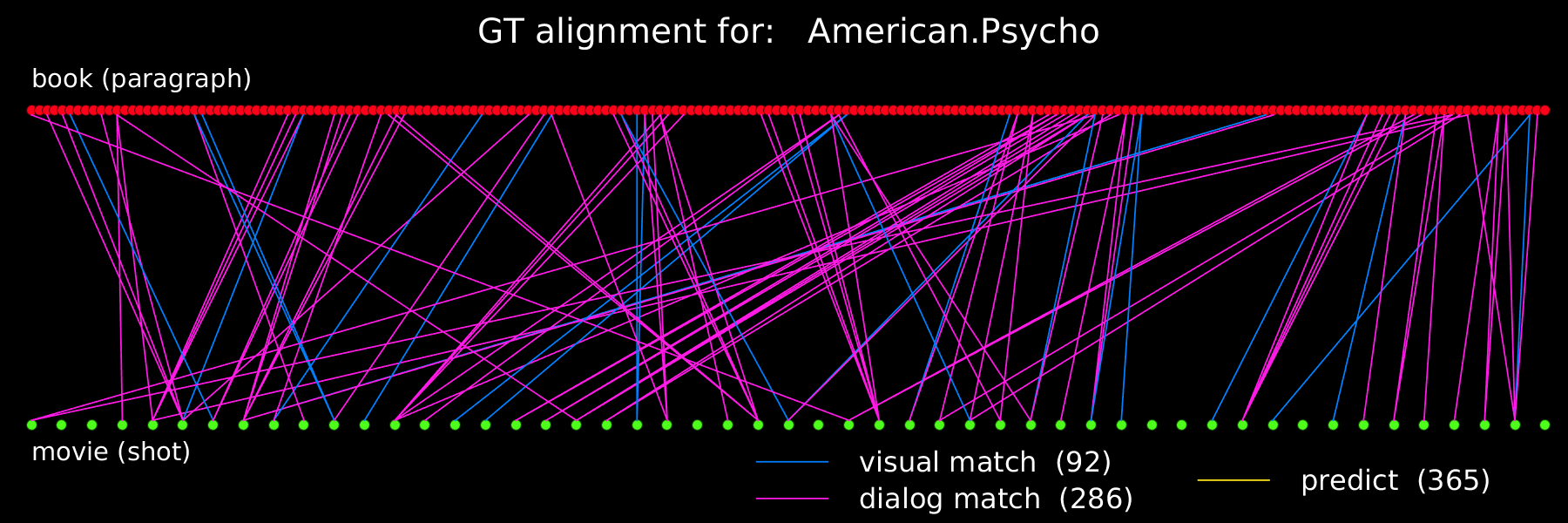}
\end{minipage}\\
\begin{minipage}{1\linewidth}
                \includegraphics[width=0.332\textwidth]{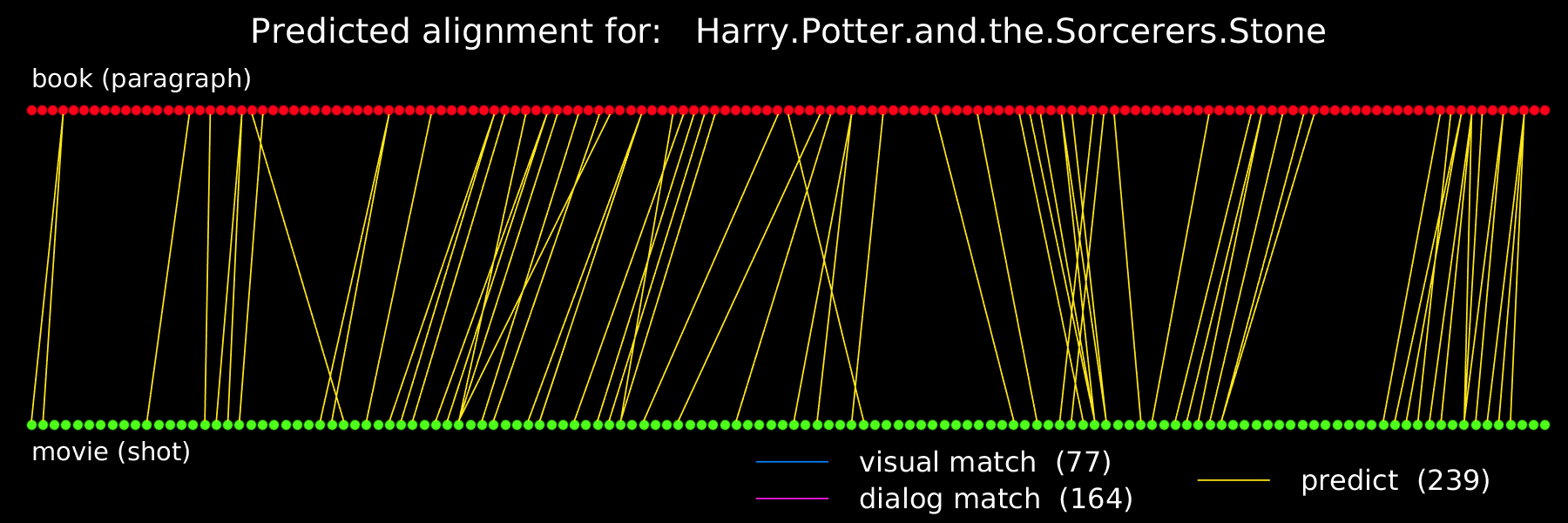}
                \includegraphics[width=0.332\textwidth]{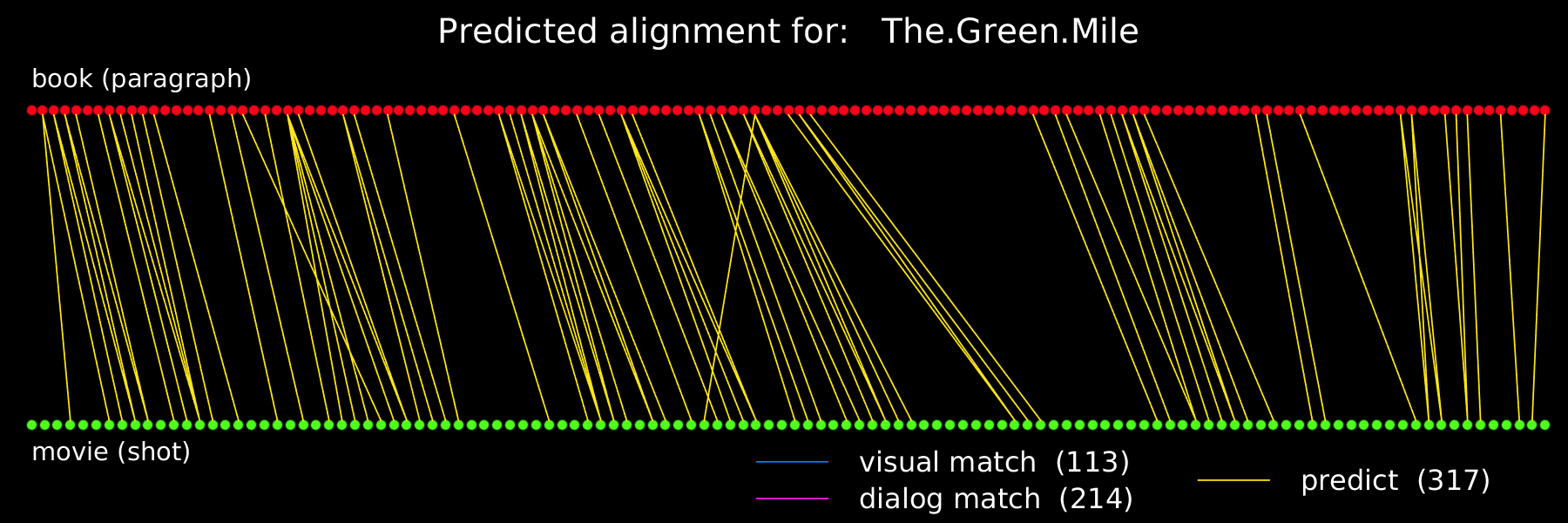}
                  \includegraphics[width=0.332\textwidth]{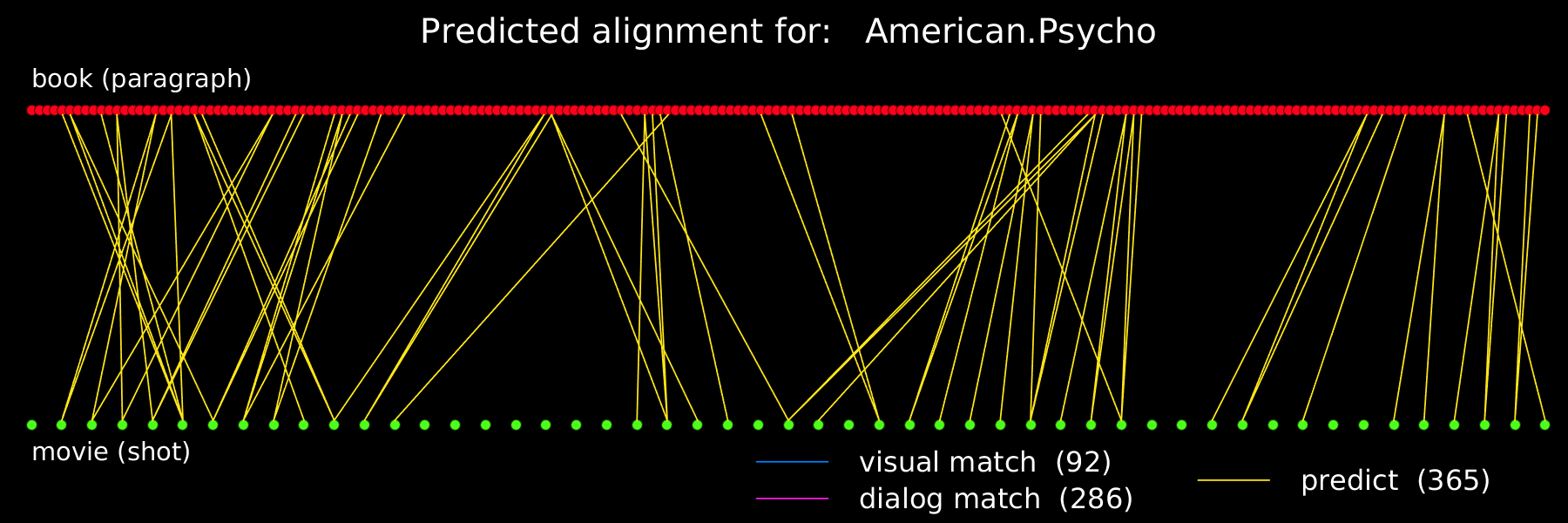}
\end{minipage}
\end{center}
\vspace{-6mm}
\caption{Alignment results of our model ({\bf bottom}) compared to ground-truth alignment ({\bf top}). In ground-truth, blue lines indicate visual matches, and magenta are the dialog matches. Yellow lines indicate predicted alignments.}\label{fig:align}
\end{figure*}

\vspace{-1mm}
\subsection{Book ``Retrieval''}
\vspace{-1mm}

In this experiment, we compute alignment between a movie and all (test) 10 books, and check whether our model retrieves the correct book. Results are shown in Table~\ref{fig:retrieval}. Under each book we show the computed similarity. In particular, we use the energy from the CRF, and scale all similarities relative to the highest one ($100$). 
Notice that our model retrieves the correct book for each movie. 

{\bf Describing a movie via other books}. We can also caption movies by matching shots to paragraphs in a corpus of books. Here we do not encourage a linear timeline (CRF) since the stories are unrelated, and we only match at the local, shot-paragraph level. We show a description for \emph{American Psycho} borrowed from the book \emph{Fight Club} in Fig.~\ref{fig:cross}. 

\vspace{-1mm}
\subsection{The CoCoBook: Writing Stories for CoCo}
\vspace{-1mm}

Our next experiment shows that our model is able to ``generate'' descriptive stories for (static) images. In particular we used the image-text embedding from~\cite{DBLP:journals/corr/KirosSZ14} and generated a simple caption for an image. We used this caption as a query, and used our sentence embedding trained on books to find top 10 nearest sentences (sampled from a few hundred thousand from BookCorpus). We re-ranked these based on the 1-gram precision of non-stop words. Given the best result, we return the sentence as well as the 2 sentences before and after it in the book. The results are in Fig.~\ref{fig:cocobook}. Our sentence embedding is able to retrieve semantically meaningful \emph{stories} to explain the images.

\renewcommand{\baselinestretch}{0.98}
\section{Conclusion}
\vspace{-1mm}
In this paper, we explored a new problem of aligning a book to its movie release. We proposed an approach that computes several similarities between shots and dialogs and the sentences in the book. We exploited our new sentence embedding in order to compute similarities between sentences. We further extended the image-text neural embeddings to  video, and proposed a context-aware alignment model that takes into account all the available similarity information. 
We showed results on a new dataset of movie/book alignments as well as several quantitative results that showcase the power and potential of our approach. 

\vspace{-1mm}
\section*{Acknowledgments}
\vspace{-1mm}
\begin{small}
We acknowledge the support from NSERC, CIFAR, Samsung, Google, and ONR-N00014-14-1-0232. We also thank Lea Jensterle for helping us with elaborate annotation, and Relu Patrascu for his help with numerous infrastructure
related problems.
\end{small}

\section*{Appendix}

\appendix
In the Appendix we provide more qualitative results.

\section{Qualitative Movie-Book Alignment Results}
  \vspace{-0.5mm}
  
 We show a few qualitative examples of alignment in Fig.~\ref{fig:align}. In this experiment, we show results obtained with our full model (CRF). For a chosen shot  (a node in the CRF) we show the corresponding paragraph in the book.
 
 We can see that some dialogs in the movies closely follow the book and thus help with the alignment. This is particularly important since the visual information is not as strong. Since the text around the dialogs typically describe the scene, the dialogs thus help us ground the visual information contained in the description and the video.

 \vspace{-1mm}
 \vspace{-0.5mm}
\section{Borrowing ``Lines'' from Other Books}
 We show a few qualitative examples of top-scoring matches for shot in a movie with a paragraph in another book (a book that does not correspond to this movie). 
 
\begin{figure*}[t!]
 \includegraphics[width=1\textwidth,trim=30 58 40 25,clip=true]{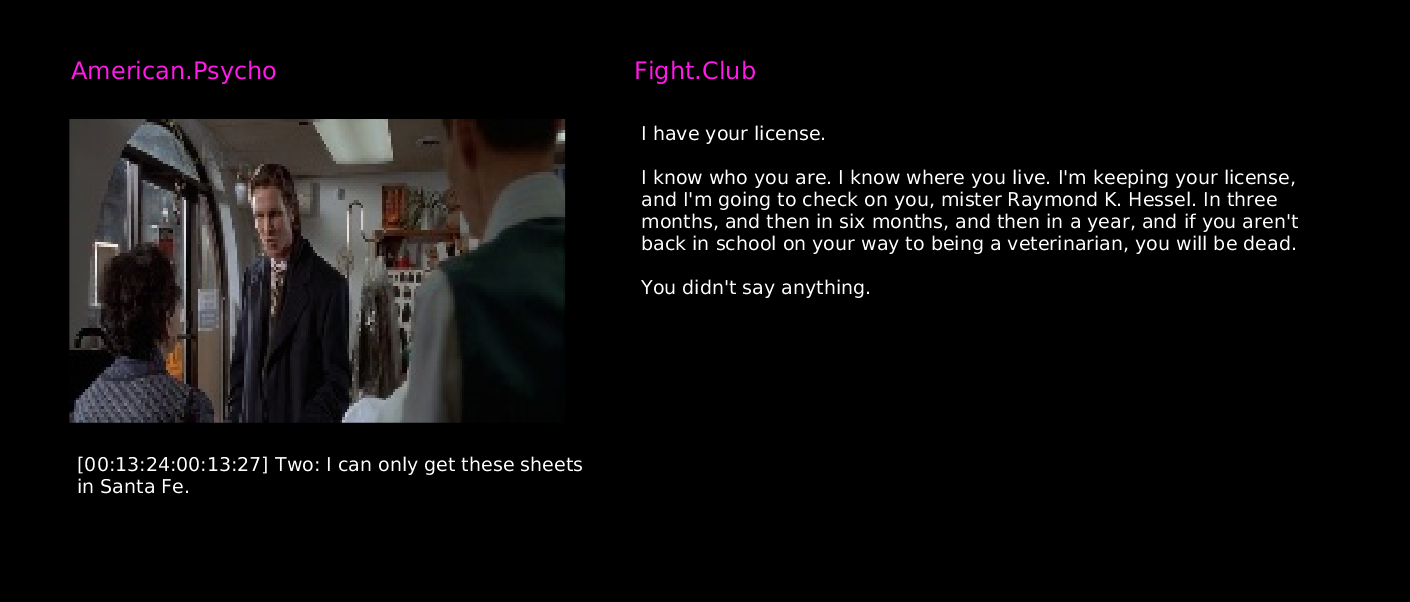}\\[2mm]
 \includegraphics[width=1\textwidth,trim=30 58 40 25,clip=true]{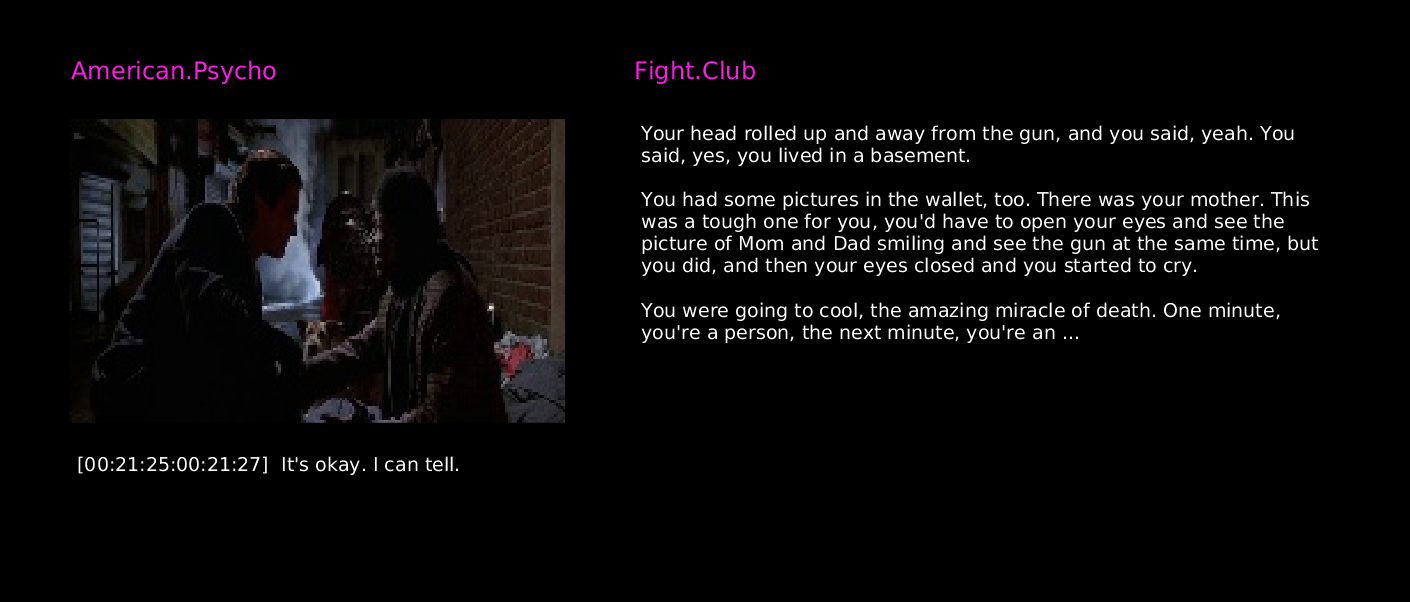}\\[2mm]
 \includegraphics[width=1\textwidth,trim=30 65 40 25,clip=true]{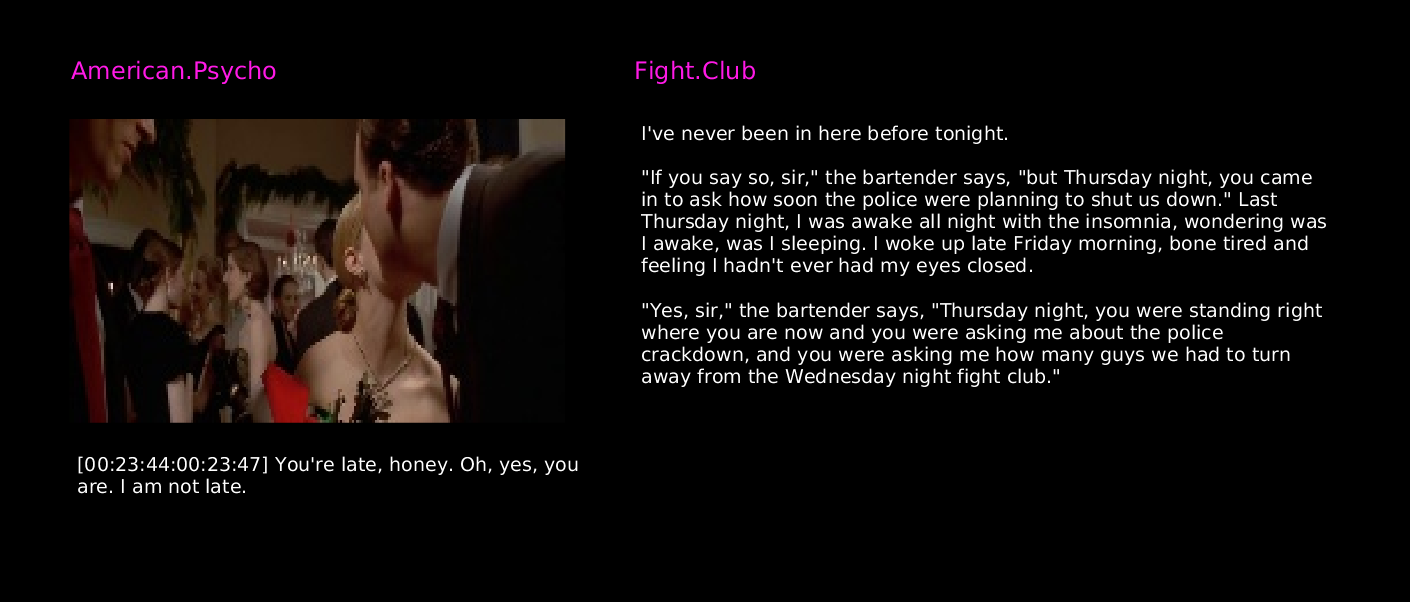}\\[-4.0mm]
 \caption{{\bf Examples of of borrowing paragraphs from other books -- {\color{magenta}{10 book experiment}}}. We show a few examples of top-scoring correspondences between a shot in a movie and a paragraph in a book that does not correspond to the movie. Note that by forcing the model to choose from another book, the top-scoring correspondences may still have a relatively low similarity. In this experiment, we did not enforce a global alignment over the full book -- we use the similarity output by our contextual CNN.}
 \label{fig:crossalign}
\end{figure*}

{\bf 10 book experiment.} In this experiment, we allow a clip in our 10 movie dataset (excluding the training movie) to match to paragraphs in the remaining 9 books (excluding the corresponding book).  The results are in Fig.~\ref{fig:crossalign}. Note that the top-scoring matches chosen from only a small set of books may not be too meaningful. 

{\bf 200 book experiment.} We scale the experiment by randomly selecting 200 books from our BookCorpus. The results are in Fig.~\ref{fig:crossalignBig}. One can see that by using many more books results in increasingly better ``stories''.

 \begin{figure*}[t!]
 \includegraphics[width=1\textwidth,trim=30 24 30 74,clip=true]{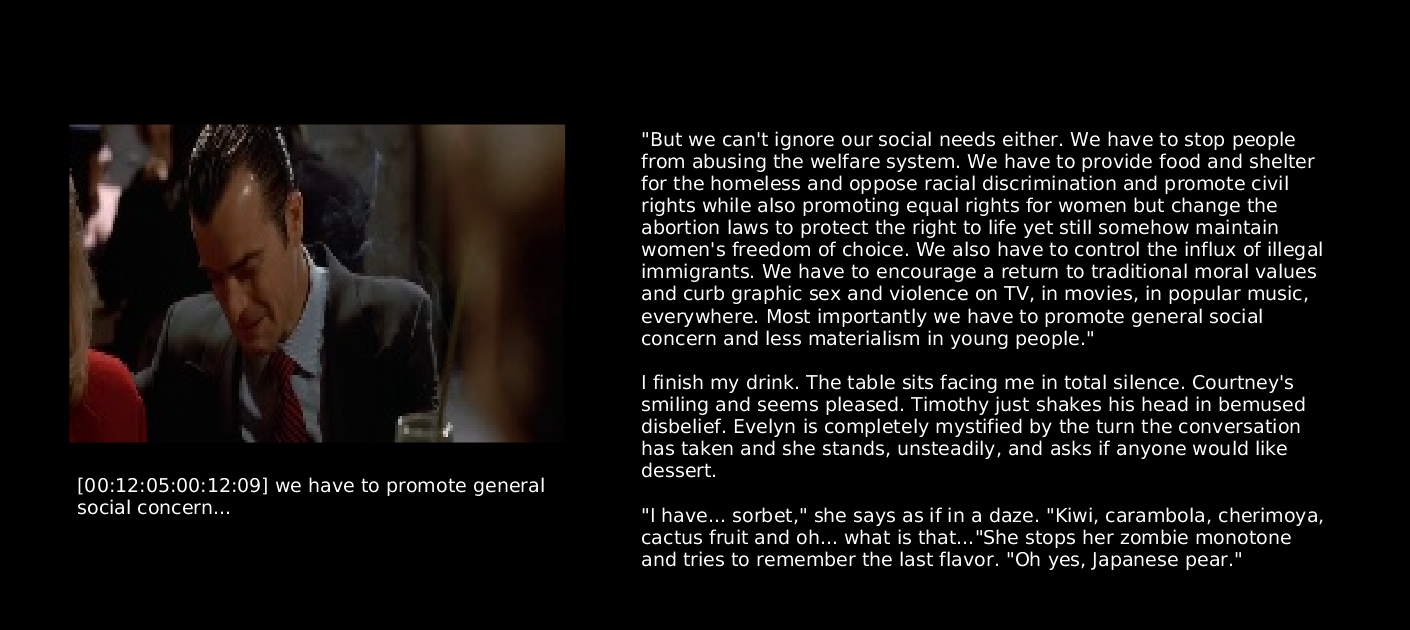}\\[-7mm]
{\color{magenta}{\hspace{3mm}American Psycho}}\\[3.5mm]
 \includegraphics[width=1\textwidth,trim=30 52 50 74,clip=true]{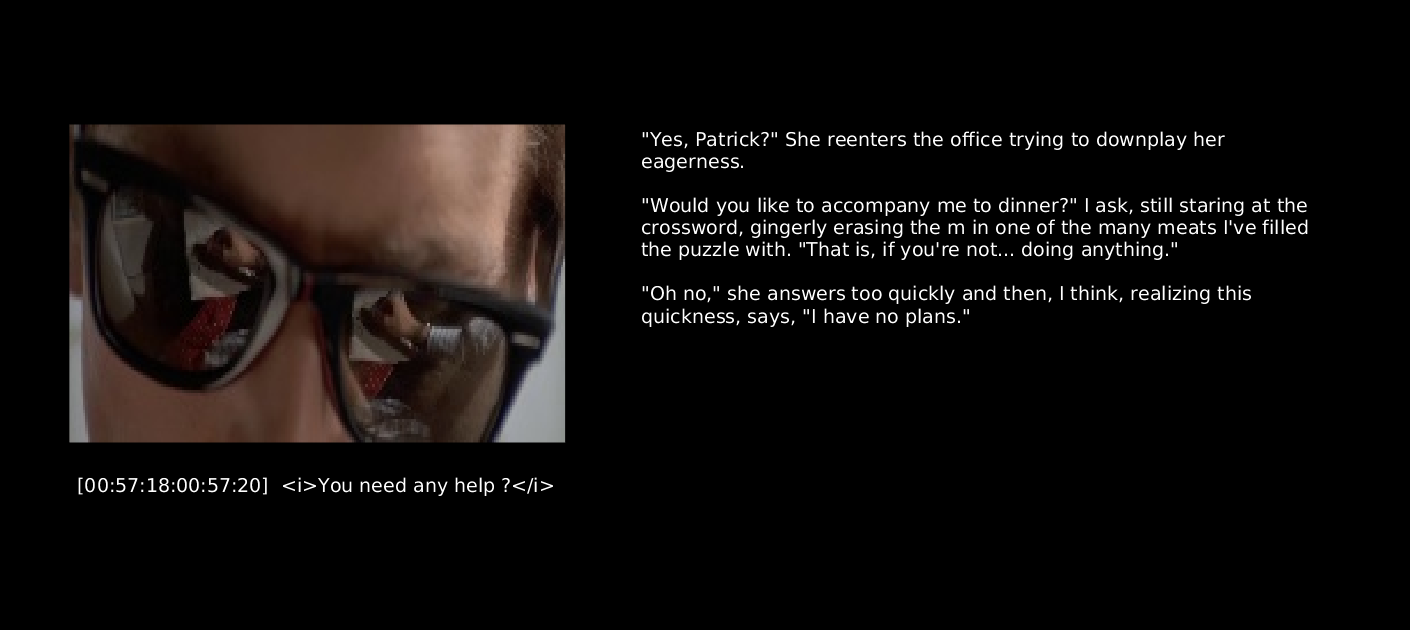}\\[-7mm]
{\color{magenta}{\hspace{3mm}American Psycho}}\\[3.5mm]
 \includegraphics[width=1\textwidth,trim=30 2 50 74,clip=true]{figs/plotAlign/HarryPotterandtheSorcerersStone_27}\\[-7mm]
{\color{magenta}{\hspace{3mm}Harry Potter}}\\[0.2mm]
 \caption{{\bf Examples of movie-book alignment}. We use our model to align a movie to a book. Then for a chosen shot (which is a node in our CRF) we show the corresponding paragraph, plus one before and one after, in the book inferred by our model. On the left we show one (central) frame from the shot along with the subtitle sentence(s) that overlap with the shot. Some dialogs in the movie closely follow the book and thus help with the alignment.}
 \label{fig:align}
\end{figure*} 

\clearpage
 \begin{figure*}[t!]
 \includegraphics[width=1\textwidth,trim=30 23 30 72,clip=true]{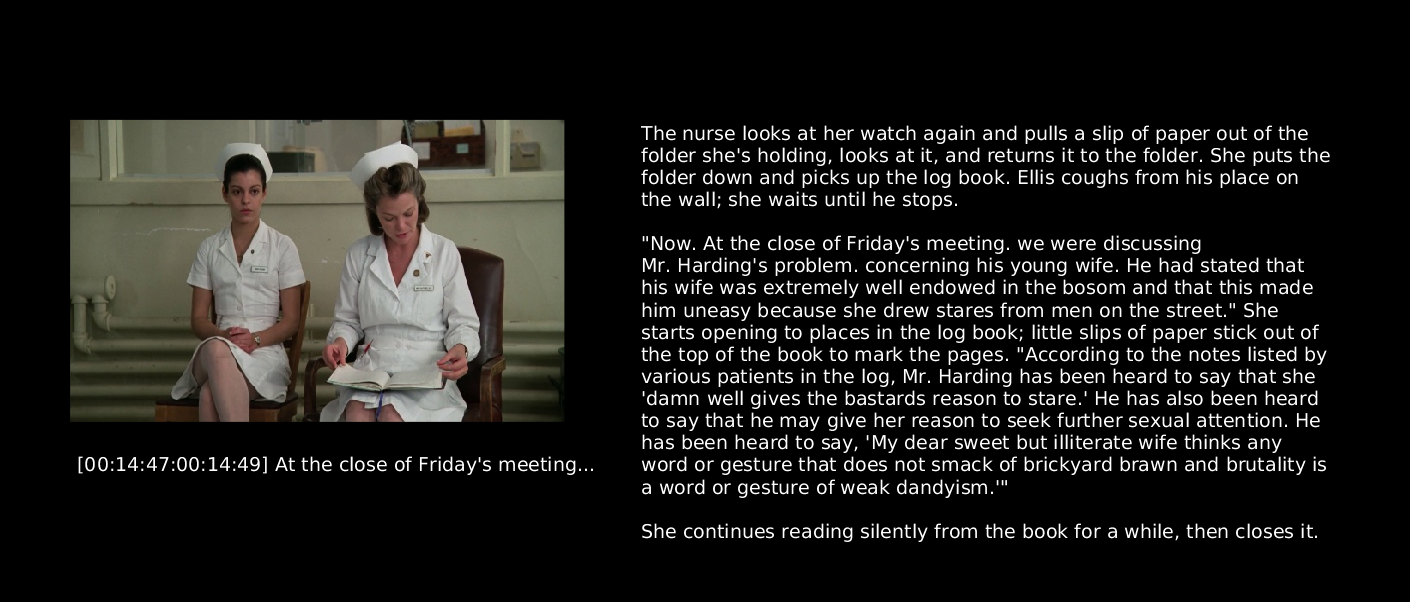}\\[-7mm]
{\color{magenta}{\hspace{3mm}One Flew Over the Cuckoo's Nest}}\\[4.4mm]
 \includegraphics[width=1\textwidth,trim=30 12 30 72,clip=true]{figs/plotAlign/OneFlewOvertheCuckooNest_28}\\[-7mm]
{\color{magenta}{\hspace{3mm}One Flew Over the Cuckoo's Nest}}\\[4.4mm]
 \includegraphics[width=1\textwidth,trim=30 26 30 72,clip=true]{figs/plotAlign/ShawshankRedemption_34}\\[-7mm]
{\color{magenta}{\hspace{3mm}Shawshank Redemption}}\\[0.2mm]
 \caption{{\bf Examples of movie-book alignment}. We use our model to align a movie to a book. Then for a chosen shot (which is a node in our CRF) we show the corresponding paragraph, plus one before and one after, in the book inferred by our model. On the left we show one (central) frame from the shot along with the subtitle sentence(s) that overlap with the shot. Some dialogs in the movie closely follow the book and thus help with the alignment.}
 \label{fig:align2}
\end{figure*} 
\clearpage

 \begin{figure*}[t!]
 \includegraphics[width=1\textwidth,trim=30 32 32 74,clip=true]{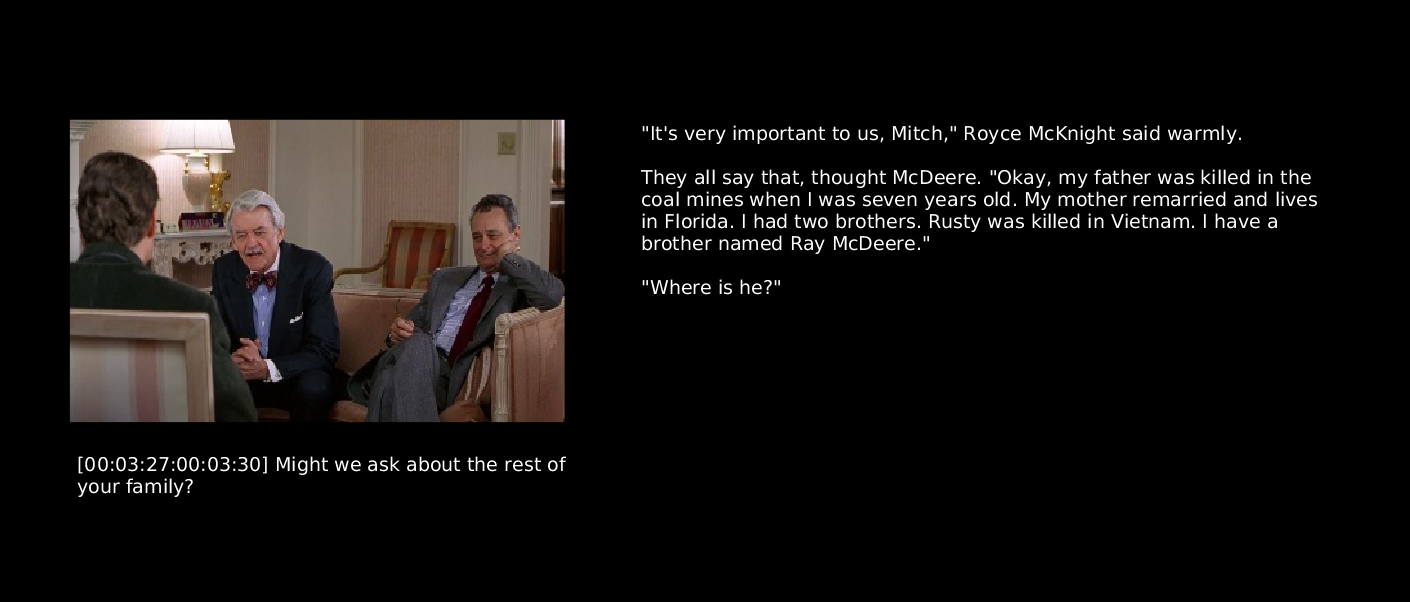}\\[-7mm]
{\color{magenta}{\hspace{3mm}The Firm}}\\[4.4mm]
 \includegraphics[width=1\textwidth,trim=30 6 30 74,clip=true]{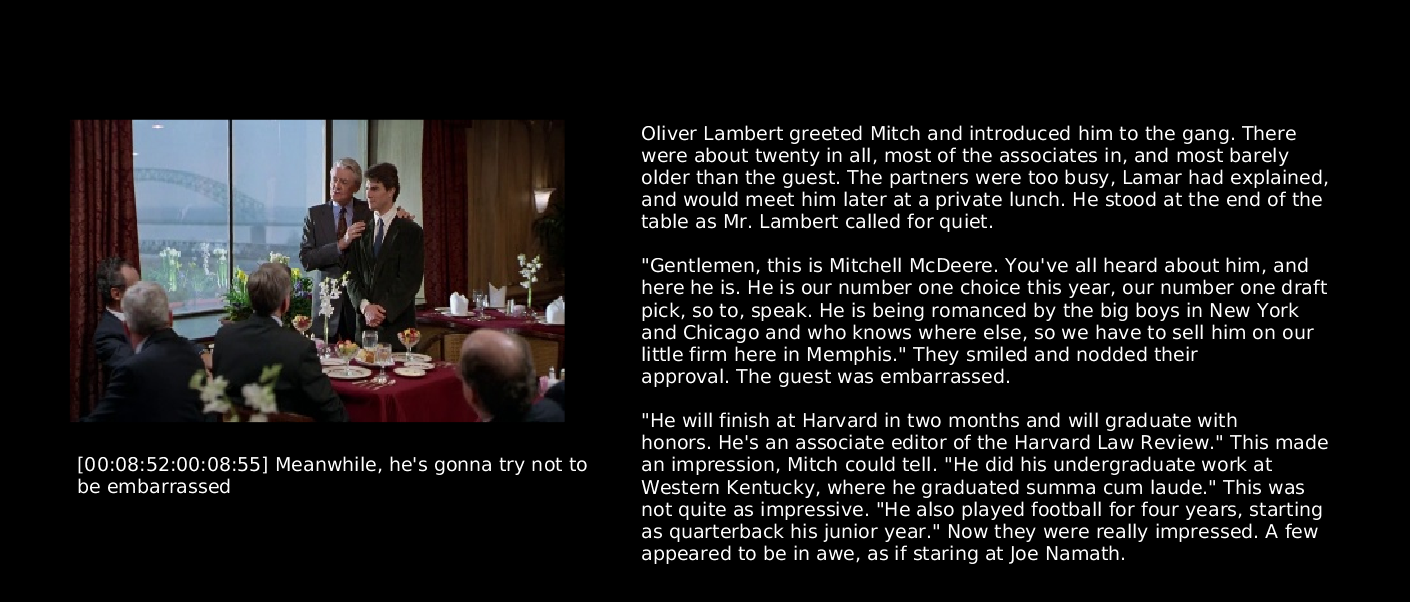}\\[-7mm]
{\color{magenta}{\hspace{3mm}The Firm}}\\[4.4mm]
 \includegraphics[width=1\textwidth,trim=30 32 30 74,clip=true]{figs/plotAlign/TheFirm_33}\\[-7mm]
{\color{magenta}{\hspace{3mm}The Firm}}\\[0.2mm]
 \caption{{\bf Examples of movie-book alignment}. We use our model to align a movie to a book. Then for a chosen shot (which is a node in our CRF) we show the corresponding paragraph, plus one before and one after, in the book inferred by our model. On the left we show one (central) frame from the shot along with the subtitle sentence(s) that overlap with the shot. Some dialogs in the movie closely follow the book and thus help with the alignment.}
 \label{fig:align3}
\end{figure*} 
\clearpage

 \begin{figure*}[t!]
 \includegraphics[width=1\textwidth,trim=30 4 32 74,clip=true]{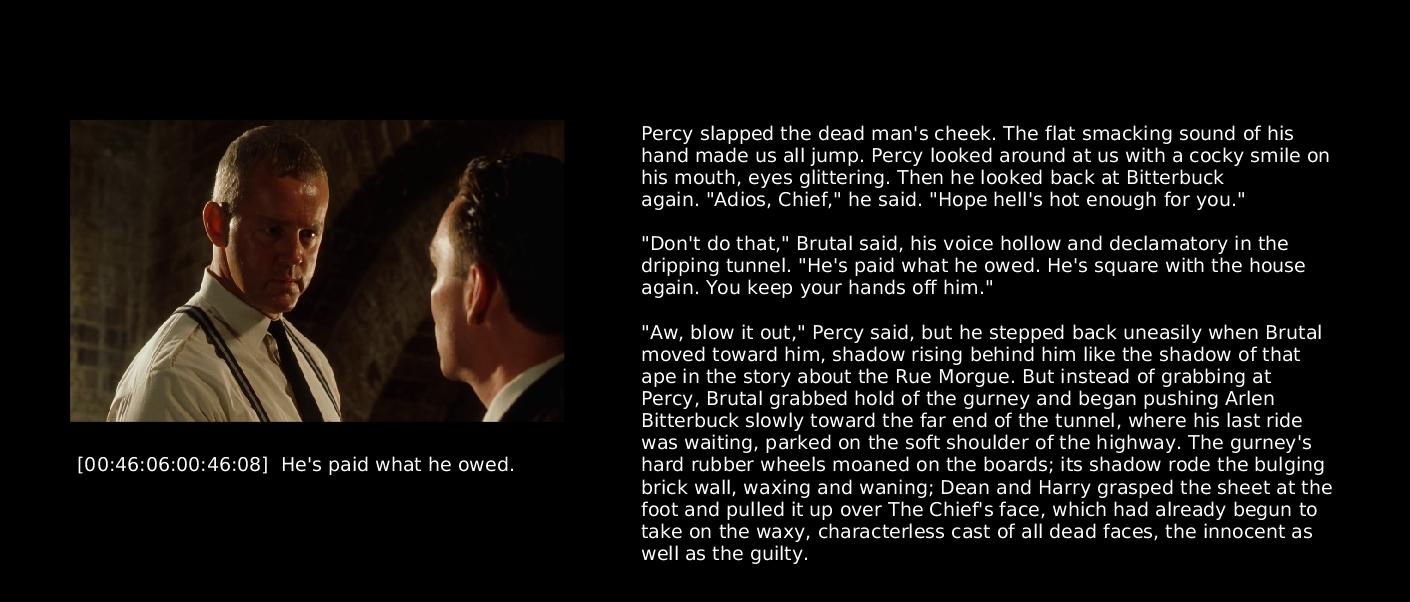}\\[-7mm]
{\color{magenta}{\hspace{3mm}The Green Mile}}\\[4.4mm]
 \includegraphics[width=1\textwidth,trim=30 42 30 74,clip=true]{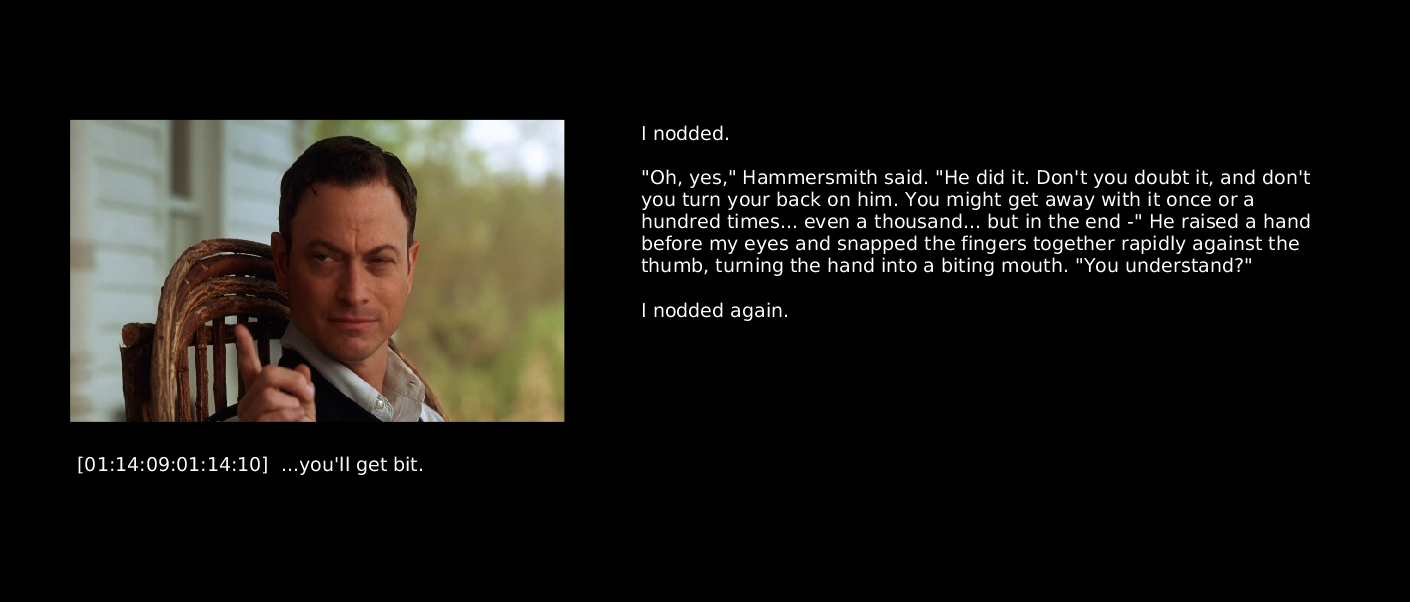}\\[-7mm]
{\color{magenta}{\hspace{3mm}The Green Mile}}\\[4.4mm]
 \includegraphics[width=1\textwidth,trim=30 42 30 74,clip=true]{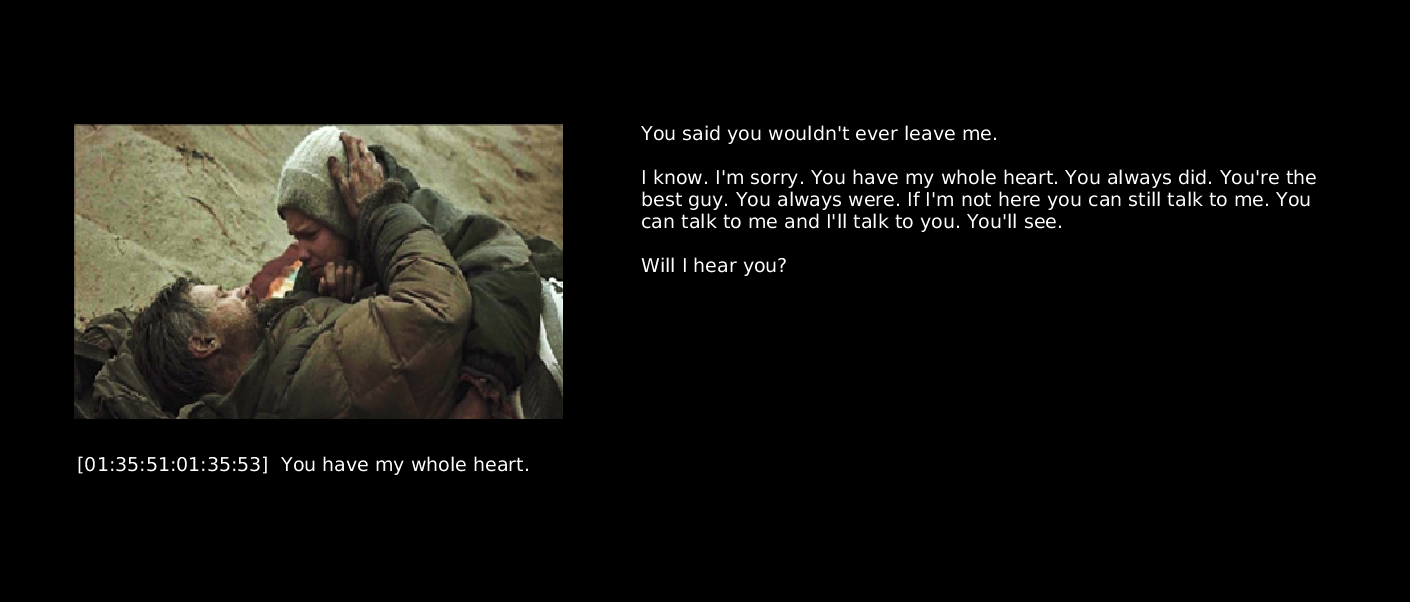}\\[-7mm]
{\color{magenta}{\hspace{3mm}The Road}}\\[0.2mm]
 \caption{{\bf Examples of movie-book alignment}. We use our model to align a movie to a book. Then for a chosen shot (which is a node in our CRF) we show the corresponding paragraph, plus one before and one after, in the book inferred by our model. On the left we show one (central) frame from the shot along with the subtitle sentence(s) that overlap with the shot. Some dialogs in the movie closely follow the book and thus help with the alignment.}
 \label{fig:align3}
\end{figure*} 

\clearpage

\clearpage

 \begin{figure*}[t!]
 \includegraphics[width=1\textwidth,trim=30 58 40 25,clip=true]{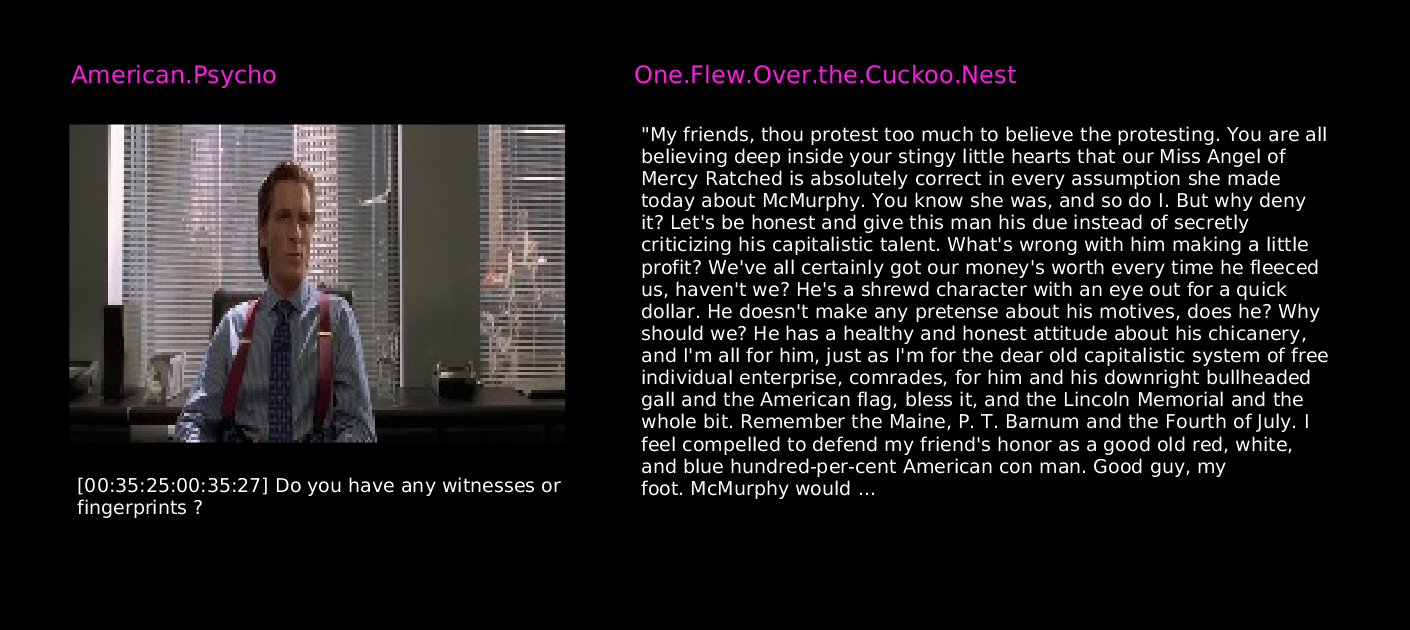}\\[2mm]
 \includegraphics[width=1\textwidth,trim=30 70 40 25,clip=true]{figs/plotX/HarryPotterandtheSorcerersStone_FightClub_1}\\[2mm]
 \includegraphics[width=1\textwidth,trim=30 55 40 25,clip=true]{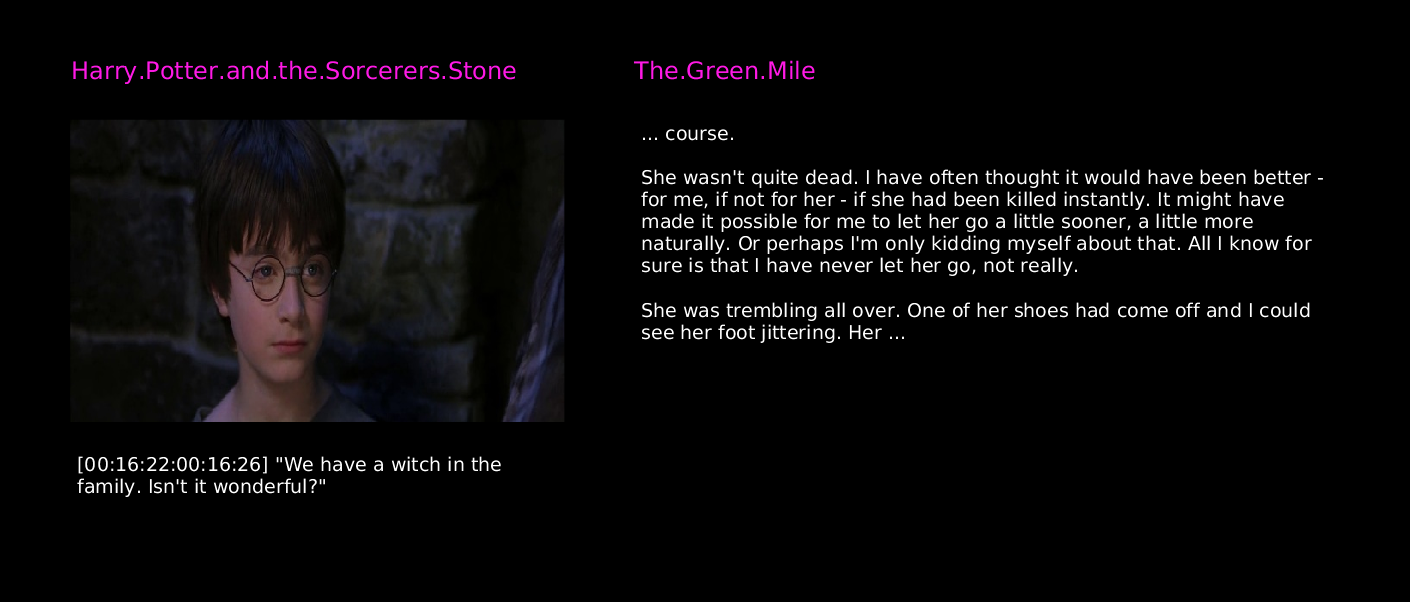}\\[-4.0mm]
 \caption{{\bf Examples of of borrowing paragraphs from other books -- {\color{magenta}{10 book experiment}}}. We show a few examples of top-scoring correspondences between a shot in a movie and a paragraph in a book that does not correspond to the movie. Note that by forcing the model to choose from another book, the top-scoring correspondences may still have a relatively low similarity. In this experiment, we did not enforce a global alignment over the full book -- we use the similarity output by our contextual CNN.}
 \label{fig:crossalign1}
\end{figure*} 

\clearpage
 \begin{figure*}[t!]
 \centering
 \includegraphics[width=1\textwidth,trim=30 58 40 25,clip=true]{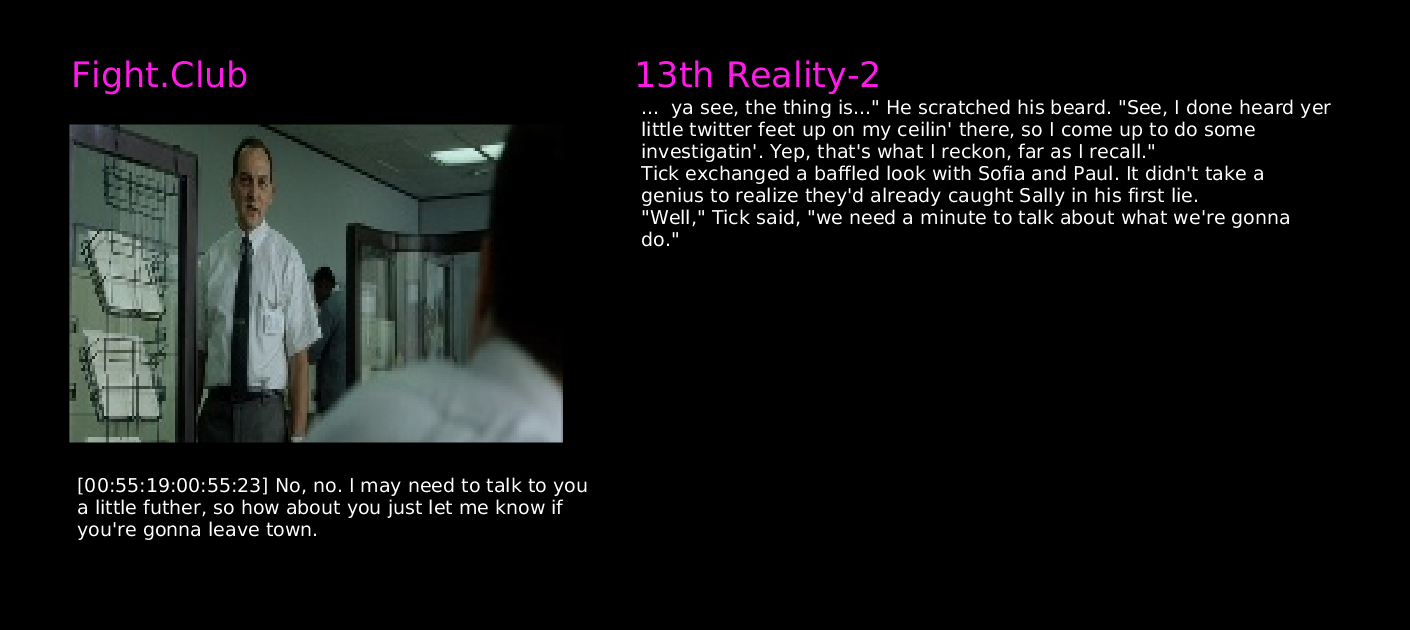}\\[2mm]
 \includegraphics[width=1\textwidth,trim=30 70 40 25,clip=true]{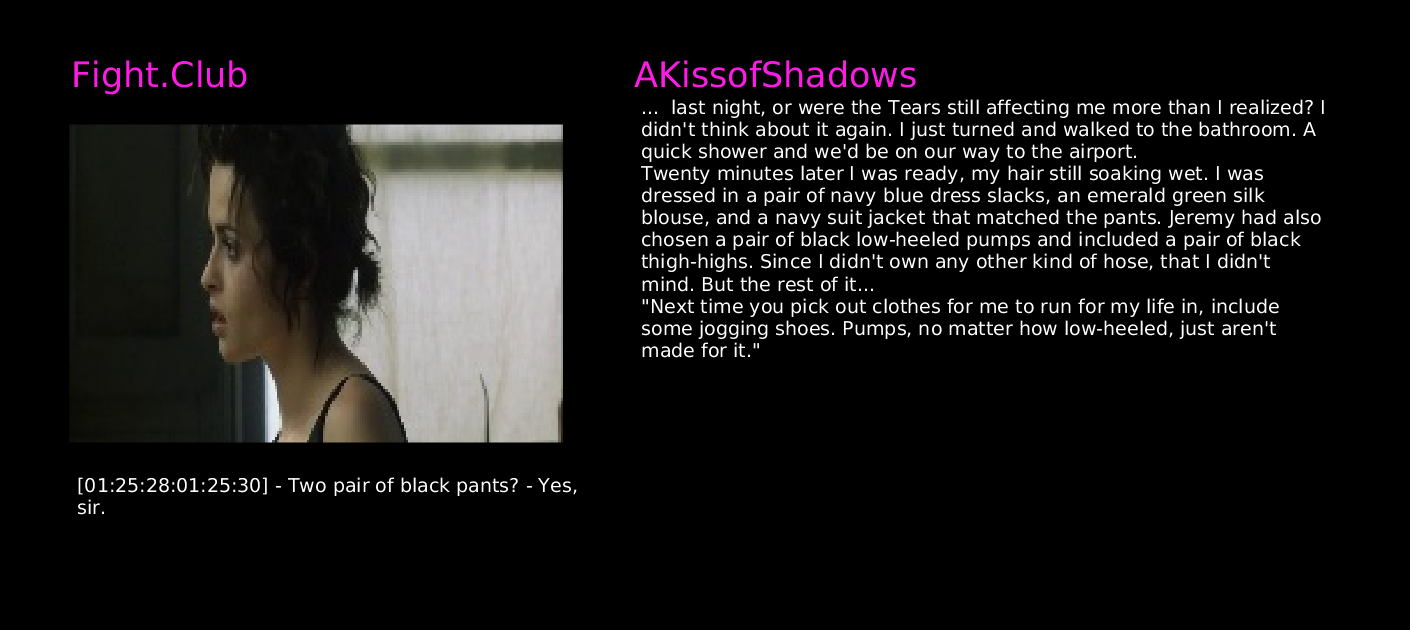}\\[2mm]
 \includegraphics[width=1\textwidth,trim=30 55 40 25,clip=true]{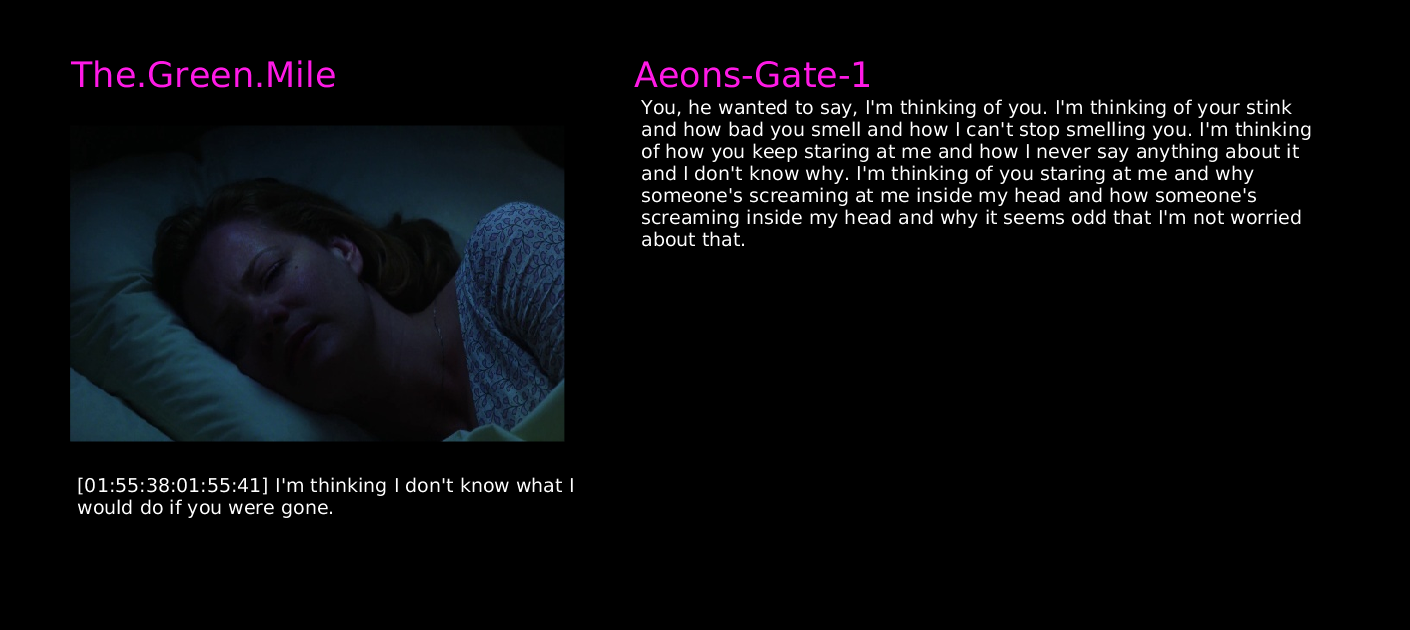}\\[-0.5mm]
 \caption{{\bf Examples of of borrowing paragraphs from other books -- {\color{magenta}{200 book experiment}}}. We show a few examples of top-scoring correspondences between a shot in a movie and a paragraph in a book that does not correspond to the movie. By scaling up the experiment (more books to choose from), our model gets increasingly more relevant ``stories''.}
 \label{fig:crossalignBig}
\end{figure*} 

\clearpage
 \begin{figure*}[t!]
 \centering
 \includegraphics[width=1\textwidth,trim=30 58 40 25,clip=true]{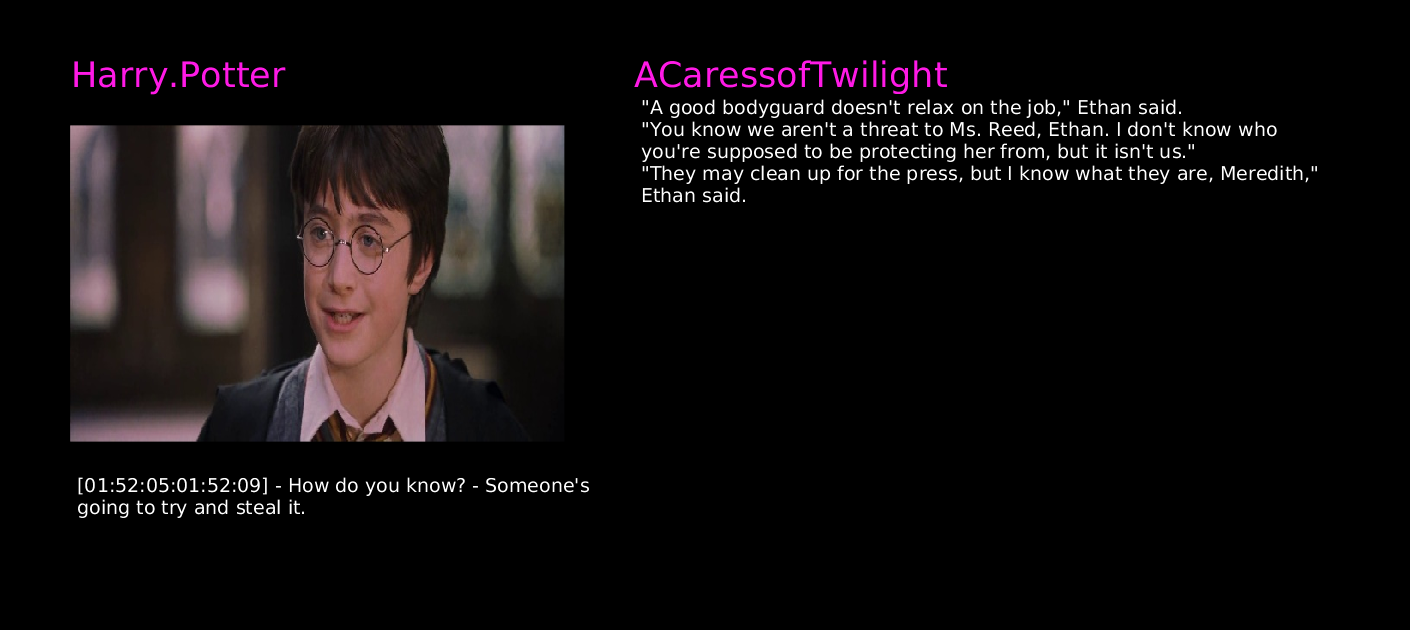}\\[2mm]
 \includegraphics[width=1\textwidth,trim=30 70 40 25,clip=true]{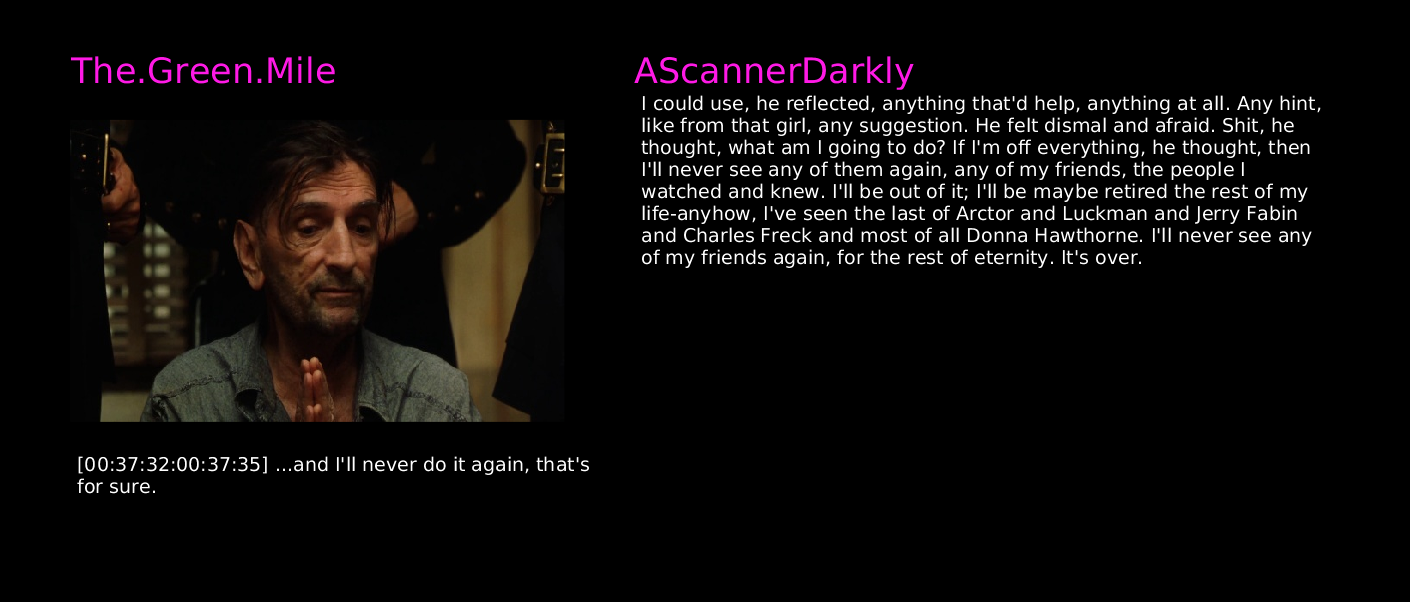}\\[2mm]
 \includegraphics[width=1\textwidth,trim=30 55 40 25,clip=true]{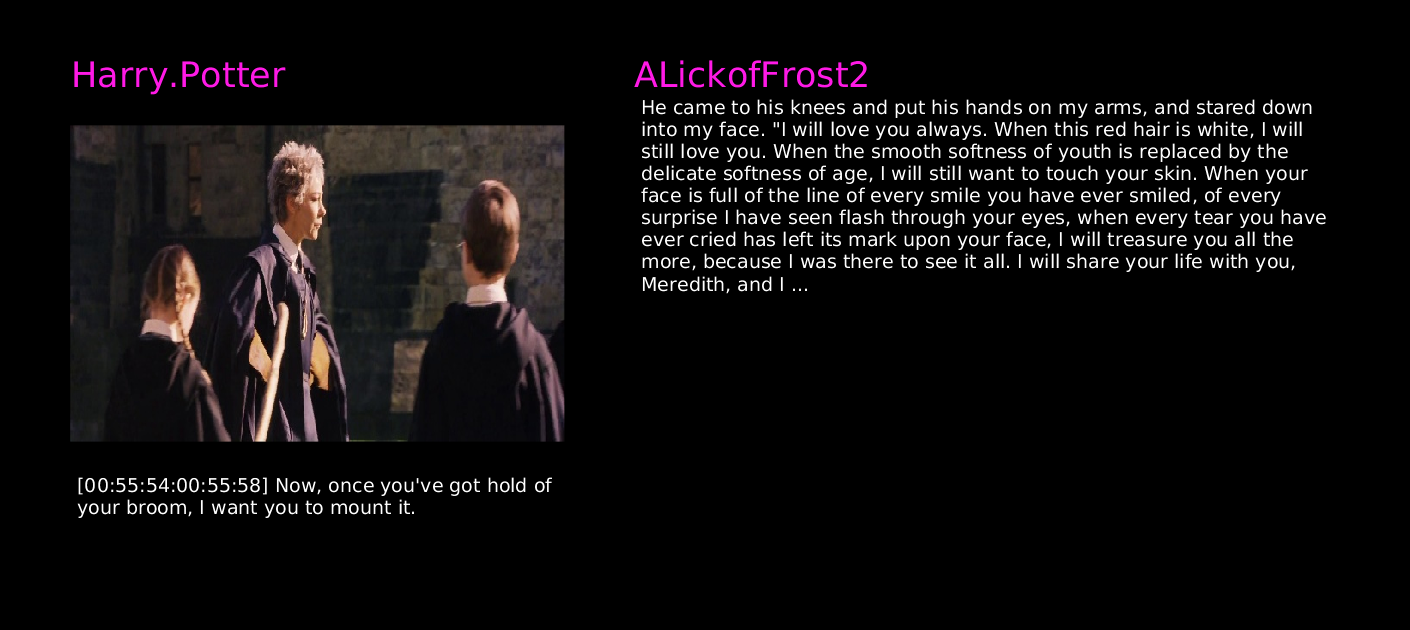}\\[-0.5mm]
 \caption{{\bf Examples of of borrowing paragraphs from other books -- {\color{magenta}{200 book experiment}}}. We show a few examples of top-scoring correspondences between a shot in a movie and a paragraph in a book that does not correspond to the movie. By scaling up the experiment (more books to choose from), our model gets increasingly more relevant ``stories''. Bottom row: failed example.}
 \label{fig:crossalignBig}
\end{figure*} 

\clearpage

\newpage
\section{The CoCoBook}

\begin{minipage}{\textwidth}
We show more results for captioning CoCo images~\cite{lin2014microsoft} with passages from the books. 
\end{minipage}

\vspace{9mm}
\begin{minipage}{\textwidth}
\begin{minipage}{0.4\textwidth}
\includegraphics[width=0.87\textwidth]{figs/laptop}
\end{minipage} 
\begin{minipage}{0.52\textwidth}
`` somewhere you 'll never find it , '' owens sneered .
if never meant five seconds , his claim was true .
the little shit 's gaze cut left , where a laptop sat on a coffee table .
trey strode to it .
owens ' email program was open .
\end{minipage}
\end{minipage}

\vspace{9mm}
\begin{minipage}{\textwidth}
\begin{minipage}{0.4\textwidth}
\includegraphics[width=0.87\textwidth]{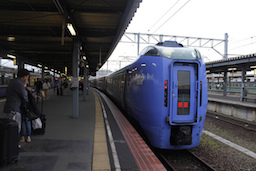}
\end{minipage} 
\begin{minipage}{0.52\textwidth}
seriously .
its like a train crashing into another train .
a train wreck .
just something like that .
i try to convince her .
\end{minipage}
\end{minipage}

\vspace{9mm}
\begin{minipage}{\textwidth}
\begin{minipage}{0.4\textwidth}
\includegraphics[width=0.87\textwidth]{figs/food}
\end{minipage} 
\begin{minipage}{0.52\textwidth}
everyone was allowed to rest for the next twenty-four hours .
that following evening : the elect , not their entourages , were called to a dining hall for supper with lady dolorous .
a table that curved inward was laden with food and drink .
the wall behind the table was windows with a view of the planet .
girls in pink stood about and at attention .
\end{minipage}
\end{minipage}

\vspace{9mm}
\begin{minipage}{\textwidth}
\begin{minipage}{0.3\textwidth}
\includegraphics[width=0.87\textwidth]{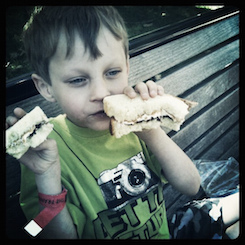}
\end{minipage} 
\begin{minipage}{0.62\textwidth}
he had simply ... healed .
brian watched his fellow passengers come aboard .
a young woman with blonde hair was walking with a little girl in dark glasses .
the little girl 's hand was on the blonde 's elbow .
the woman murmured to her charge , the girl looked immediately toward the sound of her voice , and brian understood she was blind - it was something in the gesture of the head .
\end{minipage}
\end{minipage}

\begin{figure*}[htb!]
\begin{minipage}{0.24\textwidth}
\includegraphics[width=0.87\textwidth]{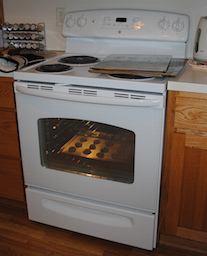}
\end{minipage} 
\begin{minipage}{0.68\textwidth}
this was a beautiful miniature reproduction of a real london town house , and when jessamine touched it , tessa saw that the front of it swung open on tiny hinges .
tessa caught her breath .
there were beautiful tiny rooms perfectly decorated with miniature furniture , everything built to scale , from the little wooden chairs with needlepoint cushions to the cast-iron stove in the kitchen .
there were small dolls , too , with china heads , and real little oil paintings on the walls .
`` this was my house . ''
\end{minipage}
\end{figure*}

\begin{figure*}[htb!]
\begin{minipage}{0.4\textwidth}
\includegraphics[width=0.87\textwidth]{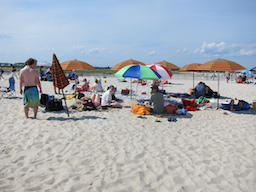}
\end{minipage} 
\begin{minipage}{0.52\textwidth}
if he had been nearby he would have dragged her out of the room by her hair and strangled her .
during lunch break she went with a group back to the encampment .
out of view of the house , under a stand of towering trees , several tents were sitting in a field of mud .
the rain the night before had washed the world , but here it had made a mess of things .
a few women fired up a camp stove and put on rice and lentils .
\end{minipage}
\end{figure*}

\begin{figure*}[htb!]
\begin{minipage}{0.4\textwidth}
\includegraphics[width=0.87\textwidth]{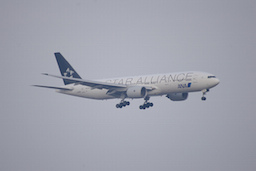}
\end{minipage} 
\begin{minipage}{0.52\textwidth}
then a frightened yell .
`` hang on ! ''
suddenly , jake was flying through the air .
nefertiti became airborne , too .
he screamed , not knowing what was happening-then he splashed into a pool of water .
\end{minipage}
\end{figure*}

\begin{figure*}[htb!]
\begin{minipage}{0.4\textwidth}
\includegraphics[width=0.87\textwidth]{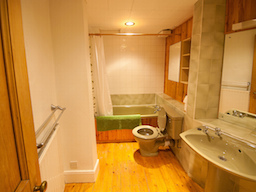}
\end{minipage} 
\begin{minipage}{0.52\textwidth}
grabbing his wristwatch off the bedside table he checked the time , grimacing when he saw that it was just after two in the afternoon .
jeanne louise should n't be up yet .
stifling a yawn , he slid out of bed and made his way to the en suite bathroom for a shower .
twenty minutes later paul was showered , dressed , and had brushed his teeth and hair .
feeling somewhat alive now , he made his way out of his and jeanne louise 's room , pausing to look in on livy as he passed .
\end{minipage}
\end{figure*}

\begin{figure*}[htb!]
\begin{minipage}{0.4\textwidth}
\includegraphics[width=0.87\textwidth]{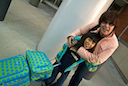}
\end{minipage} 
\begin{minipage}{0.52\textwidth}
she cried .
quentin put a heavy , warm , calming hand on her thigh , saying , `` he should be sober by then . ''
a cell phone rang .
he pulled his from his back pocket , glanced at it , then used the remote to turn the tv to the channel that showed the feed from the camera at the security gate .
`` oh , it 's rachel . ''
\end{minipage}
\end{figure*}

\begin{figure*}[htb!]
\begin{minipage}{0.25\textwidth}
\includegraphics[width=0.87\textwidth]{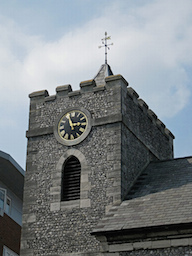}
\end{minipage} 
\begin{minipage}{0.67\textwidth}
now however she was out of his shot .
he had missed it completely until he had ended up on the ground with his shotgun .
an old clock hung on the wall near the door .
the was obviously broken , the small red hand ticking the same second away over and over again .
morgan squeezed the trigger and pellets ripped out of their package , bounced down the barrel , flew through the air and ripped into the old clock tearing it in two before it smashed to the ground .
\end{minipage}
\end{figure*}

\begin{figure*}[htb!]
\begin{minipage}{0.4\textwidth}
\includegraphics[width=0.87\textwidth]{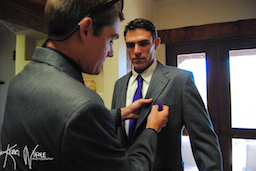}
\end{minipage} 
\begin{minipage}{0.52\textwidth}
a man sat in a chair , facing the wall opposite of me .
it nearly startled me when i first saw him , and made a bit of a squeak , but he did nothing .
he had dark gray hair , a black suit and pants , and a gray and blue striped tie .
s-sir ?
i said .
\end{minipage}
\end{figure*}

\begin{figure*}[htb!]
\begin{minipage}{0.4\textwidth}
\includegraphics[width=0.87\textwidth]{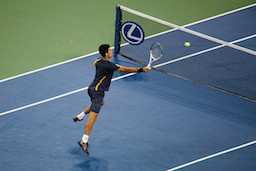}
\end{minipage} 
\begin{minipage}{0.52\textwidth}
its been years since we last played together , but as i recall , he was rather weak at the net .
or was it his serving ?
all i know is he plays tennis much better than he plays cricket .
perhaps , mr brearly , frances eventually replied , we should wait until we actually start playing .
then we can ascertain our oppositions faults , and make a plan based on the new information .
\end{minipage}
\end{figure*}

\begin{figure*}[htb!]
\begin{minipage}{0.4\textwidth}
\includegraphics[width=0.87\textwidth]{figs/vase}
\end{minipage} 
\begin{minipage}{0.52\textwidth}
since it was the middle of summer , there were candles in the fireplace instead of a fire .
but it still cast a romantic glow over the room .
there were candles on the mantle and on a table set up in the corner with flowers .
as she looked around , her eyes instinctively turned to find max who was behind a bar opening a bottle of champagne .
the doors were closed quietly behind her and her mouth felt dry as she looked across the room at the man who had haunted her dreams for so long .
\end{minipage}
\end{figure*}

\begin{figure*}[htb!]
\begin{minipage}{0.4\textwidth}
\includegraphics[width=0.87\textwidth]{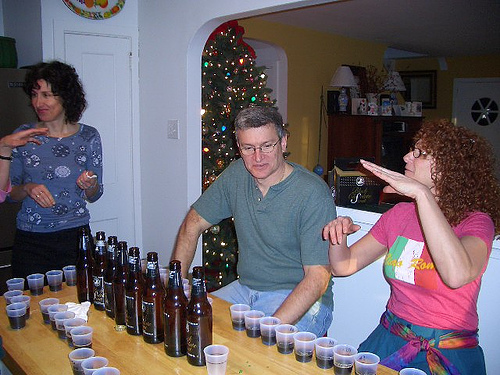}
\end{minipage} 
\begin{minipage}{0.52\textwidth}
the open doorway of another house provided a view of an ancient game of tiles .
it wasnt the game that held reddings attention .
it was the four elderly people who sat around a table playing the game .
they were well beyond their productive years and the canal township had probably been their whole lives .
redding and lin ming stepped away from the doorway right into the path of a wooden pushcart .
\end{minipage}
\end{figure*}

\begin{figure*}[htb!]
\begin{minipage}{0.4\textwidth}
\includegraphics[width=0.87\textwidth]{figs/market}
\end{minipage} 
\begin{minipage}{0.52\textwidth}
along with the fish , howard had given them some other picnic treats that had spoiled ... mushrooms in cream sauce , rotted greens .
the bats and temp were only eating from the river now , but the remaining picnic food was running low .
there were a few loaves of stale bread , some cheese , some dried vegetables , and a couple of cakes .
gregor looked over the supplies and thought about boots wailing for food and water in the jungle .
it had been unbearable .
\end{minipage}
\end{figure*}

\begin{figure*}[htb!]
\begin{minipage}{0.4\textwidth}
\includegraphics[width=0.87\textwidth]{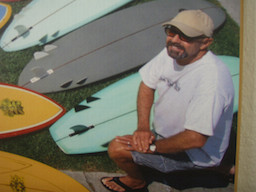}
\end{minipage} 
\begin{minipage}{0.52\textwidth}
he felt the first stirrings of fear mixing with his anger .
a light flicked on in the room and eric jerked , blinking for a minute at the brightness before the images focused .
there was a tall , thin man standing over a mannequin .
he looked like he was assembling it , since its leg was on the ground next to the man and its arm was in two pieces farther away .
then the mannequin 's head turned .
\end{minipage}
\end{figure*}

\clearpage
\balance
\vspace{-2mm}
{\small
\bibliographystyle{ieee}
\bibliography{egbib}
}

\end{document}